\pdfoutput=1
\documentclass[]{amazonpreprint}

\usepackage{mathtools}
\usepackage[textsize=tiny]{todonotes}
\usepackage{xspace}
\usepackage{booktabs}
\usepackage{marvosym}
\usepackage{graphicx}
\usepackage[table]{xcolor}
\usepackage{multirow}
\usepackage{tabularx}
\usepackage{subcaption}
\usepackage{wrapfig}
\usepackage{amsthm}
\usepackage{amsmath}
\usepackage{amsfonts}
\usepackage{amssymb}
\usepackage{enumitem}
\usepackage{adjustbox}

\theoremstyle{plain}

\theoremstyle{definition}

\theoremstyle{remark}

\newcommand{\nop}[1]{}
\newtheorem{hyp}{Hypothesis}

\newcommand{\method}{\ensuremath{\textnormal{A-Evolve}}\xspace}
\newcommand{\minhua}[1]{\textcolor{purple}{#1}}

\newcommand{\suhang}[1]{\textcolor{orange}{#1}}

\AtBeginDocument{\let\cite\citep}
\newlength{\colfigwidth}
\AtBeginDocument{\setlength{\colfigwidth}{0.75\textwidth}}
\newcommand{\frameworkfigwidth}{0.6\textwidth}
\newcommand{\panelwidth}{0.38\textwidth}

\title{Position: Agentic Evolution is\\ the Path to Evolving LLMs}

\author[1]{Minhua~Lin}
\author[2]{Hanqing~Lu}
\author[2]{Zhan~Shi}
\author[2]{Bing~He}
\author[2]{Rui~Mao}
\author[1]{Zhiwei~Zhang}
\author[1]{Zongyu~Wu}
\author[2]{Xianfeng~Tang}
\author[2]{Hui~Liu}
\author[2]{Zhenwei~Dai}
\author[3]{Rui~Zhang}
\author[1]{Xiang~Zhang}
\author[1]{Suhang~Wang}
\author[2]{Benoit~Dumoulin}
\author[4]{Jian~Pei}

\affiliation[1]{The Pennsylvania State University}
\affiliation[2]{Amazon}
\affiliation[3]{Independent Researcher}
\affiliation[4]{Duke University}

\abstract{\renewenvironment{abstract}{}{}\begin{abstract}
As Large Language Models (LLMs) move from curated training sets into open-ended real-world environments, a fundamental limitation emerges: static training cannot keep pace with continual deployment environment change. Scaling training-time and inference-time compute improves static capability but does not close this train--deploy gap. We argue that addressing this limitation requires a new scaling axis—\emph{evolution}. Existing deployment-time adaptation methods, whether parametric fine-tuning or heuristic memory accumulation, lack the strategic agency needed to diagnose failures and produce durable improvements. Our position is that \textbf{agentic evolution represents the inevitable future of LLM adaptation}, elevating evolution itself from a fixed pipeline to an autonomous evolver agent. We instantiate this vision in a \textbf{general framework}, \textsc{A-Evolve}, which treats deployment-time improvement as a deliberate, goal-directed optimization process over persistent system state. We further propose the \textbf{evolution-scaling hypothesis}: the capacity for adaptation scales with the compute allocated to evolution, positioning agentic evolution as a scalable path toward sustained, open-ended adaptation in the real world.
\end{abstract}}
\correspondence{\email{luhanqin@amazon.com}}
\metadata[Code]{\url{https://github.com/A-EVO-Lab/A-EVOLVE}}

\begin{document}
\maketitle

% The abstract is already in the title box (\abstract above).
\section{Introduction}
\label{sec:introduction}
Large Language Models (LLMs)~\cite{radford2018improving,brown2020language,touvron2023llama} have achieved remarkable progress by scaling along two primary axes: increasing training-time compute (spanning pre-training and post-training)~\cite{kaplan2020scaling, lai2025survey} and scaling inference-time compute via reasoning chains~\cite{wei2022chain, snell2025scaling}. However, real-world applications of LLMs in open-ended environments still face a fundamental challenge: the \textbf{train-deploy environment gap}~\cite{gama2014concept,hu2025testtime}. Once deployed, models trained with finite training data cannot exhaustively anticipate the infinite variety of real-world cases, shifting APIs, and evolving constraints. Therefore, purely static models inevitably degrade or fail under prolonged deployment.

This motivates the need for a new scaling axis: \emph{evolution}---the deployment-time ability of a model to autonomously improve its capabilities during interaction, without manual intervention. We define \textbf{evolution} as \emph{continual learning for LLM systems during deployment}. An LLM system’s behavior is governed by a composite policy $\pi = (\pi_\theta, \pi_S)$, where $\pi_\theta$ denotes the parametric backbone (e.g., LLM weights) and $\pi_S$ is a non-parametric \emph{persistent artifact state}, such as tools~\cite{xia2025agent0}, code~\cite{jiang2023selfevolve}, memories~\cite{chhikara2025mem0}, and structured knowledge~\cite{han2024retrieval}. Evolution corresponds to cross-episode policy improvement driven by accumulated experience:
\begin{equation}
    (\pi_{\theta}^{t+1}, \pi_{S}^{t+1}) \leftarrow F_{\mathrm{Evolve}}(\pi_{\theta}^{t}, \pi_{S}^{t}, \mathrm{Obs}[1:t]),
\end{equation}
where $\mathrm{Obs}[1:t]$ is deployment observations (e.g., interaction traces, environment feedback, rewards), and $F_{\mathrm{Evolve}}$ is the update mechanism. In this view, evolution is not a one-shot training procedure, but an \emph{ongoing process} that converts interaction evidence into lasting behavioral improvement.

\begin{figure}[t]
    \small
    \centering
        \includegraphics[width=0.85\colfigwidth]{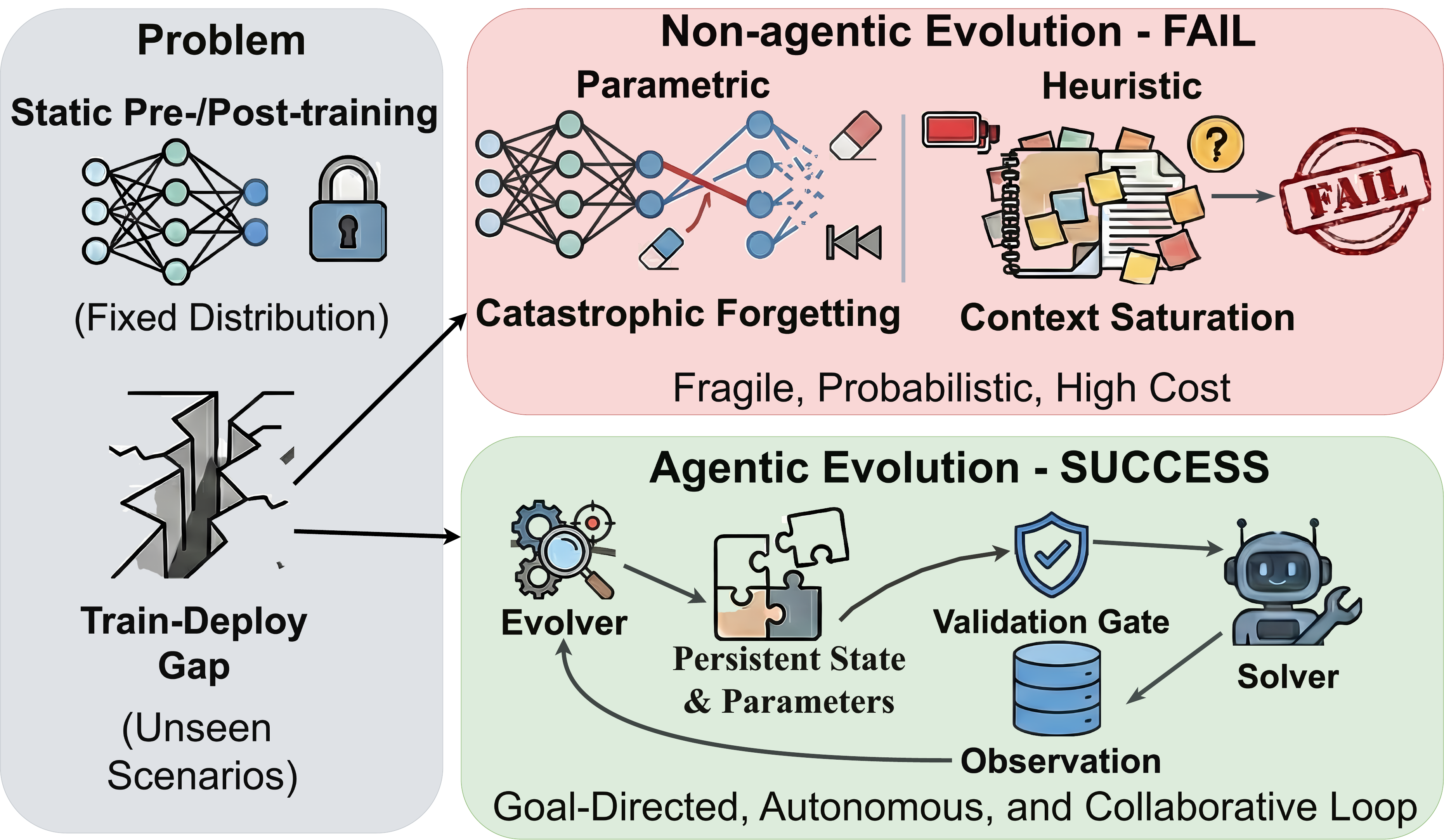}
        % \vskip -1em
    % \vskip -1.em
    \caption{From train-deploy gap to agentic evolution: An overview.}
    % \vspace{-1.5em}
    \label{fig:Framework_figure_agentic_evolution}
\end{figure}

% This observation motivates the need for a new scaling axis beyond parameters and inference compute: \emph{evolution}. Human intelligence does not rely solely on thinking harder at every encounter; it improves by accumulating experience, modifying tools, and restructuring knowledge over time. Similarly, for LLM-based agents to operate robustly in open worlds, they must possess \textbf{deployment-time adaptation}: \emph{the ability to autonomously improve their own capabilities during interaction, without relying on manual retraining}.

% We formalize \textbf{evolution} as \emph{continual learning and adapting during deployment}. An agent’s behavior is governed by a composite policy $\pi = (\pi_\theta, \pi_S)$, where $\pi_\theta$ denotes the parametric backbone (e.g., LLM weights) and $\pi_S$ denotes a non-parametric \emph{persistent artifact state}, such as tools, code, memories, and structured knowledge. Evolution corresponds to cross-episode policy improvement driven by accumulated experience:
% \[
% (\pi_{\theta}^{t+1}, \pi_{S}^{t+1}) \leftarrow F_{\mathrm{Evolve}}(\pi_{\theta}^{t}, \pi_{S}^{t}, \mathrm{Obs}[1:t]),
% \]
% where $\mathrm{Obs}[1:t]$ represents deployment observations (interaction traces, environment feedback, rewards), and $F_{\mathrm{Evolve}}$ is the update mechanism. In this view, evolution is not a one-shot training procedure, but an ongoing process that converts interaction evidence into lasting behavioral improvement.

Evolution is essential for LLM systems for at least three important reasons. First, the train-deploy gap implies that static optimization over fixed distributions is fundamentally insufficient; adaptation must occur \emph{in situ}. Second, many deployment settings impose privacy and governance constraints: user feedback, proprietary data, or sensitive interactions cannot be centrally logged for global retraining. Evolution enables \emph{local}, governed improvement through persistent artifacts without leaking private data. Third, test-time compute alone does not scale. While additional reasoning can solve novel instances, it is wasteful for recurrent failures. When a deployed LLM system repeatedly lacks a capability, such as parsing a new file format or handling a brittle API, thinking longer each time is inferior to learning once and reusing the solution. Instead, evolution amortizes expensive reasoning into cheap, persistent capability.

While several pioneering approaches to LLM evolution have emerged, they remain fundamentally limited in achieving this goal. \textbf{Parametric evolution} methods~\cite{huang2025r,hu2025test}, such as test-time training or online fine-tuning, update $\pi_\theta$ using recent observations. While expressive in principle, these updates are opaque and difficult to govern, and they risk catastrophic forgetting under distribution shift~\cite{luo2025empirical}. \textbf{Non-parametric heuristic evolution} methods~\cite{gao2025survey} instead modify $\pi_S$ by appending textual memories~\cite{wang2024agent} or optimizing prompts \emph{using fixed rules}~\cite{zhou2022large}. Although lightweight, these approaches treat evolution as a storage or search problem: as experience accumulates, memory saturates with noisy, unverified text, and improvements exhibit diminishing returns. In both cases, the update rule $F_{\mathrm{Evolve}}$ is static and heuristic, rather than adaptive and goal-directed.

These limitations are not incidental, but stem from a common root cause: the \emph{absence of agentic capability in the evolution process itself}. In existing approaches, $F_{\mathrm{Evolve}}$ is either a static optimization rule (e.g., gradient updates~\cite{xia2025agent0}) or a fixed heuristic (e.g., append-and-retrieve~\cite{shinn2023reflexion}). Such mechanisms lack the ability to \emph{reason about} failures, \emph{decide} which aspects of the system should change, and \emph{adapt} their own update strategy as deployment conditions evolve. As a result, parametric methods blindly modify weights without semantic accountability, while heuristic non-parametric methods indiscriminately accumulate experience without understanding relevance, causality, or long-term utility. Both fail precisely because evolution is treated as a mechanical procedure rather than an agentic decision-making problem.

We therefore propose \textbf{our position} in this paper: \textbf{LLM evolution has to be agentic}. The novel paradigm of \textbf{agentic evolution} is shown in  Fig.~\ref{fig:Framework_figure_agentic_evolution}. The central idea is to elevate $F_{\mathrm{Evolve}}$ from a fixed, heuristic-driven workflow to an \emph{explicit evolver agent}—a goal-directed optimizer that diagnoses what to change, autonomously governs when to change, and collaboratively synthesizes composable updates to maintain and improve both the parametric backbone $\pi_\theta$ and the persistent artifact state $\pi_S$. 

Consider a cloud platform agent deployed to compute routine metrics, such as p95 latency per endpoint, from raw production logs.
% Consider an agent deployed in a real system to compute routine metrics, such as p95 latency per endpoint, from raw production logs. 
At first, everything works smoothly—until the environment changes. A logging update renames a field, introduces a nested JSON structure, or slightly alters an interface that the agent quietly depended on. The next day, the agent starts failing. Existing evolution strategies respond in blunt ways. \emph{Parametric approaches} attempt retraining, incurring high costs and risking \emph{catastrophic forgetting}~\cite{luo2025empirical}. \emph{Non-parametric heuristic methods} instead record the failure as text, hoping the agent will re-discover the correct logic, but often face \emph{context saturation} and inconsistent retrieval. In contrast, \textbf{agentic evolution} takes a different view. The evolver treats failures as diagnostic signals, identifies \emph{what} needs to change, and uses a \emph{validation gate} to verify the fix (e.g., a regression-tested parser), ensuring the update is persistent and safe. This shift from reactive patching to deliberate, goal-directed evolution captures the essence of agentic evolution and motivates the three principles we introduce next.

% It treats such failures not as data to be absorbed or hints to be replayed, but as signals that something specific is broken. The system identifies \emph{what} capability needs to change (an outdated parsing assumption), decides \emph{when} an update is warranted rather than reacting blindly, and determines \emph{how} to fix the problem by synthesizing and validating a reusable artifact, such as a versioned log parser with regression tests. This shift—from reactive patching to deliberate capability evolution—captures the essence of agentic evolution and motivates the three principles we introduce next.

Agentic evolution is characterized by three core principles. First, the \textbf{goal-oriented principle} specifies \emph{what} to change: the evolver actively diagnoses deployment failures, attributes them to underlying causes, and targets persistent artifacts whose modification is expected to yield durable performance gains. Second, the \textbf{autonomy principle} specifies \emph{when} to change: rather than following a fixed update schedule, the evolver governs the adaptation process by selecting relevant evidence, triggering evolution only when warranted, and explicitly deciding whether to commit or reject candidate updates. Finally, the \textbf{compositional principle} specifies \emph{how} to change: improvements are realized through structured, modular artifacts—such as tools, workflows, and validation tests—produced by decomposed decision processes and integrated only after verification, while allowing periodic updates to non-parametric components when appropriate. These principles elevate evolution from a static heuristic pipeline to a deliberate, governed, and scalable decision-making process.

More broadly, agentic evolution reframes deployment-time adaptation as the joint optimization of persistent state and model parameters, rather than relying solely on the direct tuning of weights or the passive accumulation of memory. Under this view, the update rule itself becomes a \emph{budgeted optimizer}: an explicit evolver agent that decides \emph{what} to change, \emph{when} to change, and \emph{how} to change under a finite evolution-time compute budget. This leads to our central vision: \textbf{Agentic evolution represents a scalable path for the sustained, open-ended evolution of LLM systems over indefinite deployment horizons}. Unlike static heuristics that inevitably plateau, a sufficiently capable evolver can continue to extract increasingly complex improvements from accumulated experience. Continuous, open-ended evolution demands both sustainability and scalability. This motivates the \textbf{evolution-scaling hypothesis}: \emph{the capacity for adaptation—the achievable performance frontier of evolution—scales with the compute allocated to the evolution process}. We identify this as a third scaling axis, complementary to training-time and inference-time computation.
 Ultimately, adaptive intelligence in LLM systems depends not on chance, but on systematically scaling the ability to \emph{learn how to improve}.

% \suhang{does position paper add contributions??}\minhua{I didn't see this part in the prior papers}
% \section{Related Works}
% % \begin{table*}[htbp]
% %     \centering
% %     \small
% %     \caption{Systematic Analysis of Evolution Paradigms across Core Dimensions}
% %     \label{tab:evolution-comparison}
% %     \begin{tabularx}{\textwidth}{l|X|X|X|l}
% %     \toprule
% %     \textbf{Paradigm} & \textbf{Policy Updated} & \textbf{Update Mechanism ($F_{\text{Evolve}}$)} & \textbf{Meta-Scaling} & \textbf{Analogy} \\ \midrule
    
% %     \textbf{Offline Parametric Evolution} & $\pi_\theta$ (Offline) & Offline Training Loop (None at test time) & \textbf{None}: Static during deployment. & Exam Crammers \\ \hline
    
% %     \textbf{Online Parametric Evolution} & $\pi_\theta$ (Online) & Gradient Descent / Adapter Tuning & \textbf{Unstable}: Prone to forgetting & Brain Surgery \\ \hline
    
% %     \textbf{Non-parametric Heuristic-based Evolution} & $\pi_S$ (Text) & Fixed Heuristic (Append/Search) & \textbf{Low}: Context Saturation & Cluttered Note-takers \\ \hline
    
% %     \textbf{Agentic Evolution} & $\pi_S$ (Code/Tools) & \textbf{Evolver Agent} (Policy-driven) & \textbf{High}: Scales with Evolver Capability & \textbf{Autonomous Developer} \\
% %     \bottomrule
% %     \end{tabularx}
% % \end{table*}

% \subsection{From Training-time Update to Deployment-time Adaption}
% \noindent\textbf{Trianing-time update.}

% \noindent\textbf{Test-time scaling.}

% \noindent\textbf{Deployment-time Adaption.}

% \subsection{Evolution in LLMs}
% \noindent\textbf{Self-play}
% xxx

% % \subsection{Tool Learning on LLMs}

\section{Agentic Evolution and Principles}
\label{sec:agentic_evolution_setion}

We consider a deployed \emph{LLM system} that repeatedly interacts with an environment over a sequence of episodes. At episode $t$, the system observes an input or state $x_t$, produces actions $a_t$, and receives feedback $o_t$ from the environment. We model the system as a \emph{composite policy}
\begin{equation}
    \pi_t = (\pi_{\theta,t}, \pi_{S,t}),
\end{equation}
where $\pi_{\theta,t}$ denotes the parametric backbone (e.g., an LLM model with parameters $\theta$), and $\pi_{S,t}$ denotes a persistent, editable artifact state that conditions behavior across episodes, such as tools, skills, workflows, structured knowledge, and validation assets. Crucially, $\pi_{t}$ persists beyond a single interaction and constitutes the system’s deployment-time interface to capability.

\emph{Evolution} refers to cross-episode improvement that occurs \emph{during deployment}. Rather than treating deployment as a static inference phase, evolution formalizes the process by which a system converts accumulated interaction evidence into lasting behavioral change. Let $\mathrm{Obs}_{1:t}$ denote the deployment evidence collected up to episode $t$, including trajectories, tool traces, errors, and feedback. Evolution is defined as a cross-episode update process:
\begin{equation}
(\pi_{\theta,t+1}, \pi_{S,t+1})
\leftarrow
F_{\mathrm{Evolve}}(\pi_{\theta,t}, \pi_{S,t}, \mathrm{Obs}_{1:t}),
\label{eq:evolution_update_rewrite}
\end{equation}
where $F_{\mathrm{Evolve}}$ is an update mechanism that may modify the parametric model $\pi_\theta$, the persistent state $\pi_S$, or both. This definition abstracts over a broad spectrum of approaches, ranging from offline training and online fine-tuning to heuristic memory updates. Detailed taxonomy is in Appendix~\ref{appendix:full_details_evolution_category}.

\emph{Agentic evolution} instantiates $F_{\mathrm{Evolve}}$ as an explicit \emph{evolver agent}: a goal-directed, autonomous optimizer that reasons over accumulated deployment evidence and produces \emph{persistent, governed} updates. Rather than executing a fixed script, the evolver treats evolution itself as a decision-making problem. At each episode $t$, it proposes a structured candidate update $\Delta_t$, such as add, patch, refactor, or prune operations over $\pi_S$ and/or $\pi_\theta$, and makes an explicit commit decision:
\begin{equation}
\label{eq:agentic_evolution_generation}
\Delta_t \leftarrow F_{\mathrm{Evolve}}(\pi_{\theta,t}, \pi_{S,t}, \mathrm{Obs}_{1:t}), \;
c_t \leftarrow \mathcal{C}(\pi_t, \Delta_t, \mathrm{Obs}_{1:t}),
\end{equation}
followed by
\begin{equation}
\pi_{t+1} \leftarrow
\begin{cases}
\mathrm{Apply}(\pi_t, \Delta_t) & \text{if } c_t = 1, \\
\pi_t & \text{if } c_t = 0,
\end{cases}
\end{equation}
where $c_t \in \{0, 1\}$ is the commit decision from the governance gate $\mathcal{C}$. This gate is typically instantiated via automated verification (e.g., unit tests, regression checks) or human-in-the-loop review. The defining characteristic of agentic evolution is that updates are treated as \emph{conditional decisions}—explicitly proposed, evaluated, and either committed or rejected—rather than as unconditional steps in a fixed workflow.

Agentic evolution is governed by three core principles that specify \emph{what} to evolve, \emph{when} to evolve, and \emph{how} evolution is carried out.

The \textbf{goal-oriented principle} specifies \emph{what} to change. The evolver does not blindly accumulate experience or optimize generic proxies; instead, it explicitly diagnoses deployment failures, attributes them to actionable causes, and targets specific components of the persistent state whose modification is expected to improve \emph{future} performance. This shifts adaptation from correlation-based updates to causal, capability-level repair. In practice, goal orientation is realized by localizing failures to missing tools, brittle logic, interface mismatches, or incomplete workflows, and then applying targeted edits to the corresponding artifacts in $\pi_S$ (and, when appropriate, to $\pi_\theta$). By making the adaptation objective explicit and forward-looking, the system converts raw experience into durable improvements rather than transient fixes.

The \textbf{autonomy principle} specifies \emph{when} to change. Instead of following a pre-defined update schedule (e.g., evolve after every failure), the evolver controls the update decision process itself. It selects which evidence is relevant, determines whether a failure is actionable or merely transient noise, and explicitly decides whether to commit an update or perform a no-op. This decision-theoretic control allows evolution compute to be allocated only when improvement is feasible and worthwhile. Autonomy distinguishes agentic evolution from non-agentic pipelines in which updates are implicitly triggered or hard-coded, and it is essential for stable long-horizon deployment in open-ended environments.

The \textbf{compositional principle} specifies \emph{how} to evolve. Internally, the evolver is naturally decomposed into cooperating decision functions—such as diagnosis, planning, updating, and verification—that operate over shared evidence and state. This decomposition improves reliability and expressivity, whether realized within a single system or as a multi-agent workflow. Externally, evolution produces \emph{modular, structured artifacts}—for example, executable tools, reusable workflows, and validation tests—rather than unstructured text. This compositionality enables \emph{amortization}: recurring reasoning and fragile deliberation are compiled into reusable capabilities, allowing the system to bypass context saturation and avoid the diminishing returns typical of heuristic memory accumulation.
% \hanqing{I feel the compositional principle doesn't neccessarily refer to diag-plan-update loop, but refer to we could apply modular changes to different artifacts}

These principles elevate evolution from a static, heuristic-driven procedure to a governed and scalable decision-making process. By treating deployment-time improvement as an agentic optimization problem over persistent state, agentic evolution provides a foundation for LLM systems that can reliably adapt to open-ended environments while maintaining stability, interpretability, and scalability.

To illustrate, let us revisit the cloud-log agent scenario from Sec.~\ref{sec:introduction}. Let $\pi_t = (\pi_{\theta,t}, \pi_{S,t})$ denote the composite policy. Under the schema drift scenario (e.g., nested JSON changes), parametric evolution attempts to update the backbone $\pi_{\theta}$ via fine-tuning, entangling a localized interface mismatch with global model behavior and risking catastrophic forgetting. Heuristic methods instead append raw traces to $\pi_{S}$, forcing the solver to repeatedly regenerate brittle logic at inference time with limited reuse or guarantees.

Agentic evolution reframes such failures as a cross-episode optimization problem. Deployment trajectories are aggregated into structured evidence, from which an explicit evolver diagnoses the underlying cause, such as an outdated schema assumption, and formulates a targeted update objective. Rather than modifying weights or appending raw memories, the evolver proposes precise, structured edits to $\pi_S$, such as synthesizing a versioned adapter function to permanently handle the new nested format and rectifying the API schema definition. Candidate updates are evaluated and committed conditionally through explicit verification, ensuring that only validated improvements persist. In this way, agentic evolution amortizes repeated inference-time reasoning into durable, governed capability, transforming recurrent failures into stable system assets.

\section{A-Evolve: A General Framework for Agentic Evolution of LLM Systems}
\label{sec:method}

\begin{figure}[t]
    \small
    \centering
    \includegraphics[width=\frameworkfigwidth]{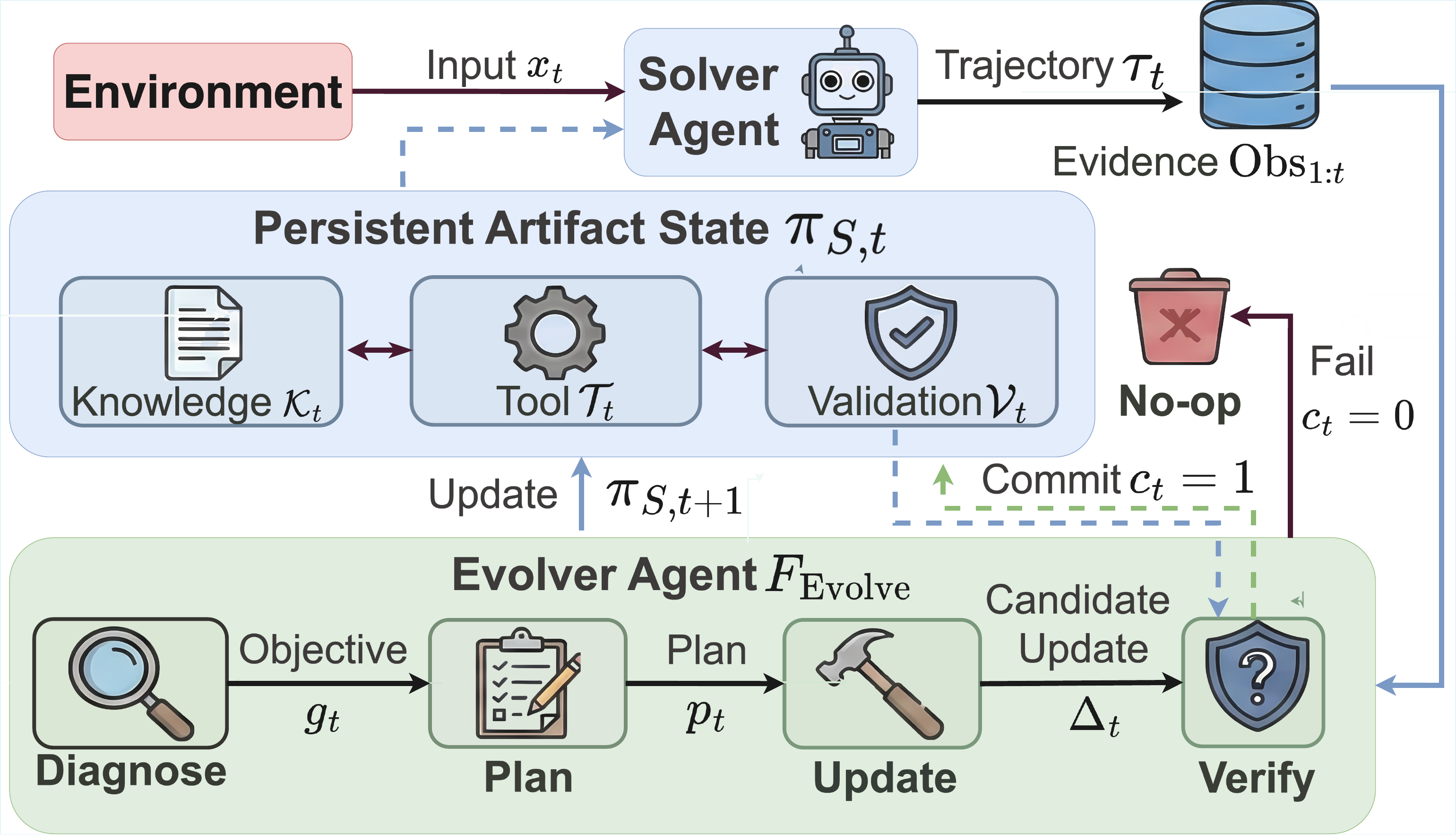}
    % \vspace{-0.5em}
    \caption{Framework of A-Evolve.}
    % \vspace{-1.5em}
    \label{fig:Framework_figure_a_evolve}
\end{figure}

A-Evolve is a general, implementation-agnostic framework that instantiates the principles of agentic evolution introduced in Sec.~\ref{sec:agentic_evolution_setion}. As illustrated in Fig.~\ref{fig:Framework_figure_a_evolve}, A-Evolve makes deployment-time evolution explicit, governed, and amortizable, enabling scalable capability improvement beyond static inference.

A-Evolve is built around three commitments. First, evolution is \emph{goal-oriented}: from deployment evidence, the system derives an explicit update objective $g$ and proposes targeted edits $\Delta$ over a structured space. Second, evolution is \emph{autonomous}: updates are conditional, driven by explicit evidence selection and a commit/no-op decision $c$, rather than a fixed schedule. Third, evolution is \emph{compositional}: the update rule $F_{\mathrm{Evolve}}$ is realized as a modular evolver that produces typed artifacts and commits them only through explicit acceptance mechanisms such as validation or review. We instantiate these commitments through three structural components: a persistent artifact state, a solve--evolve control loop, and an explicit evolver.

\paragraph{Persistent Artifact State \(\pi_S\).}
A-Evolve elevates the non-parametric component $\pi_S$ to a first-class, editable interface to system capability. Instead of accumulating transient context, deployment experience is compiled into persistent artifacts that can be created, revised, validated, and reused across episodes. This exposes a concrete edit space for evolution, allowing targeted updates to specific components rather than diffuse prompt modification. At episode $t$, the artifact state is
\begin{equation}
    \pi_{S,t} = \{\mathcal{K}_t, \mathcal{T}_t, \mathcal{V}_t\}.
\end{equation}

The \textbf{knowledge registry} $\mathcal{K}_t$ stores structured or textual artifacts such as schemas, workflows, interface contracts, and exemplars. Artifacts are addressable and versioned, enabling retrieval, patching, and replacement. This supports goal-oriented evolution by allowing failures to be localized to specific knowledge components.

The \textbf{tool registry} $\mathcal{T}_t$ contains executable functions, including scripts and API wrappers, with explicit input--output signatures and associated tests. During solve-time, tools provide deterministic action primitives that reduce inference variance. During evolve-time, they serve as diagnostic instruments for replaying failures, probing edge cases, and extracting structured error signals.

The \textbf{validation registry} $\mathcal{V}_t$ contains governance assets such as unit tests, regression suites, and optional human review hooks. Validation artifacts are themselves editable. Crucially, $\mathcal{V}_t$ grounds the commit decision $c_t$: updates are committed only if they pass verification, preventing regressions and uncontrolled drift over long deployment horizons.

\paragraph{The Solve--Evolve Loop.}
A-Evolve explicitly separates instance-level task execution from cross-episode capability improvement, ensuring that evolution is a governed process with its own control flow and compute budget. For episode $t$ with input $x_t$, the solve phase produces a trajectory
\begin{equation}
    \tau_t = \text{Solve}(\pi_t, x_t),
\end{equation}
which is appended to the cumulative evidence buffer $\mathrm{Obs}_{1:t} = \mathrm{Obs}_{1:t-1} \cup \{\tau_t\}$. The evolve phase then produces a conditional update
\begin{equation}
    \pi_{t+1} = \pi_t \oplus (c_t \cdot \Delta_t),
\end{equation}
where $\Delta_t$ and $c_t$ are defined below.

In the \textbf{solve phase}, the system executes under the composite policy $\pi_t=(\pi_{\theta,t},\pi_{S,t})$, retrieving artifacts from $\mathcal{K}_t$ and invoking tools from $\mathcal{T}_t$ as needed. The artifact state $\pi_{S,t}$ is treated as read-only to ensure reproducibility.

In the \textbf{evolve phase}, triggered asynchronously or when sufficient evidence accumulates, the evolver analyzes $\mathrm{Obs}_{1:t}$ and proposes a structured update $\Delta_t$ together with a commit decision $c_t$. The evolver may invoke tools from $\mathcal{T}_t$ to reproduce failures or analyze traces. Candidate updates are evaluated against $\mathcal{V}_t$; only validated updates are committed. This separation ensures that solve-time compute targets task completion, while evolve-time compute targets amortized improvement across episodes.

\paragraph{The Evolver \(F_{\mathrm{Evolve}}\).}
A-Evolve implements $F_{\mathrm{Evolve}}$ as an explicit evolver composed of four cooperating functions:
\begin{equation}
\small
\begin{aligned}
&g_t \leftarrow \text{Diagnose}(\mathrm{Obs}_{1:t}, \pi_{S,t}), \quad p_t \leftarrow \text{Plan}(g_t, \pi_{S,t}), \\
& \Delta_t \leftarrow \text{Update}(p_t, \pi_{S,t}), \quad 
c_t \leftarrow \text{Verify}(\Delta_t, \mathcal{V}_t).
\end{aligned}
\end{equation}

\textbf{Diagnose} identifies actionable failure modes and their causes (e.g., missing schemas or brittle tools), optionally invoking $\mathcal{T}_t$ to extract structured error signatures, and produces an update objective $g_t$.  
\textbf{Plan} translates $g_t$ into an explicit edit plan specifying target artifacts, edit operators (add, patch, refactor, prune), and ordering constraints.  
\textbf{Update} executes the plan by synthesizing concrete artifact changes, assembling a candidate update $\Delta_t$ with provenance and tests.  
\textbf{Verify} evaluates $\Delta_t$ against $\mathcal{V}_t$ and returns a commit decision $c_t \in \{0,1\}$, allowing the evolver to reject harmful or brittle updates.

\paragraph{Edit Operators and Auditing.}
A-Evolve supports a small set of canonical edit operators over $\pi_S$, including adding or patching tools, schemas, and tests, and pruning obsolete artifacts. For example, under schema drift, the evolver may add a new schema, patch a parser tool, and introduce regression tests, committing the update only if all checks pass. All proposed updates are logged with full provenance, enabling auditing, rollback, and risk-based human oversight.

\paragraph{Summary.}
A-Evolve provides a concrete framework for agentic evolution. By exposing a typed persistent artifact state, separating solve-time execution from evolve-time optimization, and enforcing modular, verifiable updates, A-Evolve enables LLM systems to convert deployment experience into durable, governed capability improvements without sacrificing stability or interpretability.

\section{The Evolution-Scaling Hypothesis}

A central motivation for agentic evolution is to enable \emph{autonomy}: the ability of an LLM system to adapt reliably in open-ended, non-stationary environments where increased training-time compute~\cite{kaplan2020scaling,lai2025survey} and increased test-time compute~\cite{schaeffer2025how} inevitably fall short. While existing evolution-style methods~\cite{cai2025training,ouyang2025reasoningbank} demonstrate that deployment-time adaptation is possible, they are typically driven by ad-hoc heuristics and fixed update rules. As a result, their effectiveness is often fragile, difficult to predict, and prone to early saturation.

This leads to a fundamental question for long-horizon deployment and, more broadly, for the path toward AGI:
\emph{Can evolution itself be made scalable and sustainable?}  
Specifically, if we allocate more resources to the evolution process—such as stronger evolvers, more analysis steps, or larger evolution-time compute budgets—does the effectiveness and speed of adaptation increase in a predictable and systematic way, or is improvement largely bounded by chance and problem-specific heuristics?

\paragraph{The Evolution-Scaling Hypothesis.}
We propose the \textbf{Evolution-Scaling Hypothesis}, which frames deployment-time adaptation not as a stochastic or opportunistic phenomenon, but as a scalable optimization process governed by resources. Analogous to established scaling laws for pre-training and inference, the hypothesis posits that allocating additional compute to the \emph{evolution loop}—rather than only to solving individual tasks—systematically improves the attainable level of adaptation.

Concretely, we argue that evolution-time compute enables an LLM system to (i) diagnose failures more accurately, (ii) consider more candidate updates, (iii) synthesize more robust artifacts, and (iv) apply stronger verification before committing changes. These capabilities compound over episodes because validated updates persist. As a result, evolution is not merely a collection of local fixes, but a convergent process whose effectiveness scales with the resources devoted to it. This perspective shifts the paradigm from hoping that heuristics generalize to \emph{engineering adaptation through compute}.

\paragraph{Evolution-Scaling Formalism.}
To make this notion precise, we define a compute-optimal view of evolution. Let $P(\pi)$ denote the performance of a deployed policy $\pi$ under a fixed environment and a fixed solve-time compute budget. Given an initial policy $\pi_0$, we define the \emph{compute-optimal evolution frontier} as
\begin{equation}
\small
P^*(C_{\mathrm{evolve}}, \pi_0)
\;\triangleq\;
\max_{F_{\mathrm{evolve}}}
\;\mathbb{E}_{\pi \sim F_{\mathrm{evolve}}(\pi_0)}
\big[ P(\pi) \big],
\end{equation}
where $F_{\mathrm{evolve}}$ ranges over all evolution strategies whose total evolution-time compute cost does not exceed $C_{\mathrm{evolve}}$. Here, $C_{\mathrm{evolve}}$ captures the resources allocated to the evolution process, including analysis steps, tool invocations, candidate synthesis, and verification, while implicitly reflecting the capability of the evolver itself.

The Evolution-Scaling Hypothesis states that, holding the environment and solve-time compute fixed, the frontier $P^*$ is strictly increasing with respect to evolution-time compute. Formally, for any
$C_{\mathrm{evolve}}^{(1)} < C_{\mathrm{evolve}}^{(2)}$, we hypothesize:
\begin{equation}
\small
P^*(C_{\mathrm{evolve}}^{(1)}, \pi_0)
\;<\;
P^*(C_{\mathrm{evolve}}^{(2)}, \pi_0).
\end{equation}
% \suhang{I think $P^*(C_{\mathrm{evolve}}^{(1)}, \pi_0)
% \;\le\;
% P^*(C_{\mathrm{evolve}}^{(2)}, \pi_0)$ might be more convincing} 
Intuitively, allocating more evolution-time compute enlarges the space of feasible update strategies, enabling deeper diagnosis, more reliable edits, and stronger governance, and thereby raising the achievable performance ceiling.

\paragraph{Interpretation.}
This hypothesis reframes deployment-time learning as a \emph{predictable scaling regime}. Rather than viewing adaptation as a series of isolated, heuristic-driven repairs, evolution is treated as an optimization process whose convergence rate and asymptotic performance depend on the strength and compute budget of the evolver. In this view, insufficient evolution resources lead to noisy, brittle improvements, while sufficient resources guarantee systematic progress toward the compute-optimal frontier.

\paragraph{Strategic Implications.}
The evolution-scaling perspective yields two complementary research directions:

\emph{Approaching the frontier.}
For a fixed evolution-time budget, the goal is to design evolution algorithms $F_{\mathrm{evolve}}$ that efficiently approach $P^*$. Agentic evolution advances this direction by replacing fixed heuristics with autonomous, structured decision-making, as empirically validated in Sec.~\ref{sec:evaluation_effectiveness}.

\emph{Raising the frontier.}
Beyond algorithmic efficiency, the hypothesis predicts that aggressively scaling evolution-time resources, such as evolver capability and analysis depth, systematically raises the performance ceiling itself. In long-horizon deployment, reallocating compute from repeatedly ``thinking harder'' at solve time to ``evolving better'' across episodes converts raw compute into durable capability, enabling mastery of increasingly complex and shifting environments. We will empirically observe the evolution scaling effect in Sec.~\ref{sec:meta_scaling_analysis}. 

\section{Empirical Studies}
\label{sec:evaluation}

In this section, we empirically evaluate the feasibility and advantages of agentic evolution by addressing three questions:
(i) \textbf{Effectiveness:} does agentic evolution improve task performance under \emph{compute-matched} conditions?
(ii) \textbf{Feasibility:} how do the core components of agentic evolution contribute to performance gains and loop reliability?
(iii) \textbf{Evolution-scaling:} does evolution capacity increase with evolution-time compute and evolver capability? 

% \bing{Since we highlight the evolver agent works under a budget-constraint limit. We can add more tests on different budgets in the future. We have tested the budget when the number of steps is lower than 30. We can measure the token cost as well. This can help.}

\subsection{Experimental Setup}
\label{sec:evaluation_setup}

\noindent\textbf{Datasets.}
We evaluate on AppWorld~\cite{trivedi2024appworld}, a widely used benchmark for tool-using agents with executable environments and unit tests for goal verification. We sample $50$ tasks from the training split for evolution and $50$ tasks from the test-normal split for evaluation. Dataset details and the sampling protocol are in Appendix~\ref{appenidx:dataset_details}.

\noindent\textbf{Implementation Details.}
We implement A-Evolve as a two-level system consisting of a \emph{solver}, which executes tasks, and an \emph{evolver}, which accumulates persistent improvements across episodes. The evolver decomposes the evolve step into four roles: diagnoser, planner, updater, and verifier, which cooperate through a shared evolution state. Unless otherwise specified, we use Claude Sonnet 4.5~\cite{anthropic2025claudesonnet45} as the evolver backbone, and evaluate solvers ranging from Claude Haiku 4.5~\cite{anthropic2025claudehaiku45} and Claude Sonnet 4/4.5 to GPT-5~\cite{singh2025openai} and Gemini 3 Flash~\cite{google2025gemini3flash}. Additional details are in Appendix~\ref{appendix:a_evolve_implementation}.

\noindent\textbf{Evaluation Protocol.}
Following prior work~\cite{trivedi2024appworld,cao2025remember}, we report two metrics: \emph{Task Goal Completion} (TGC), the fraction of tasks successfully completed, and \emph{Average Passed Tests} (APT), the average fraction of unit tests passed per task. To ensure fair comparison, we fix a solve-time compute budget $C_{\mathrm{solve}}$ (maximum tool calls, steps, or tokens per task) and an evolve-time compute budget $C_{\mathrm{evolve}}$ (maximum tokens and tool invocations per episode) across all methods. Full evaluation details are given in Appendix~\ref{appendix:evaluation_metrics}.

\begin{table*}[t]
\centering
% \small
\caption{Comparison of agentic evolution against baselines on AppWorld. We report TGC and APT (\%) across solver backbones.\vspace{-2mm}}
\label{tab:effectiveness}
\begin{tabular}{lcc|cc|cc|cc|cc}
\toprule
& \multicolumn{2}{c}{\textbf{Claude Haiku 4.5}} &  \multicolumn{2}{c}{\textbf{Claude Sonnet 4.5}} &  \multicolumn{2}{c}{\textbf{Claude Sonnet 4}}  &  \multicolumn{2}{c}{\textbf{GPT 5}} &  \multicolumn{2}{c}{\textbf{Gemini 3 Flash}}   \\
\cmidrule{2-11} 
\textbf{Method} & TGC & APT & TGC & APT & TGC & APT & TGC & APT & TGC & APT \\
\midrule
Vanilla & 32 & 51.16 & 86 & 94.94 & 42 & 79.42 & 80 & 94.15 & 56 & 80.45 \\
APE & 30 & 56.00 & 80 & 85.00 & 44 & 82.00 & 82 & 95.00 & 52 & 84.00 \\
AWM & 46 & 65.76 & 88 & \textbf{97.32} & 50 & 86.29 &84 & 94.08 & 52 & 87.75\\
A-Evolve & \textbf{64} & \textbf{84.31} & \textbf{90} & 96.47 & \textbf{56} & \textbf{88.21} & \textbf{88} & \textbf{97.02} & \textbf{82} & \textbf{92.05} \\
\bottomrule
\end{tabular}
% \vspace{-2mm}
\end{table*}

\subsection{Effectiveness: Agentic vs. Non-agentic Evolution}
\label{sec:evaluation_effectiveness}

\noindent\textbf{Setting.}
We compare \method against a \textit{non-evolving} reference and two representative \textit{non-agentic evolution} baselines and:
(i) \textit{APE}~\cite{zhou2022large}, a search-based prompt evolution method that proposes candidate instructions and selects among them via task-level scoring;
(ii) \textit{AWM}~\cite{wang2024agent}, an experience-based workflow memory baseline that induces reusable workflows from past trajectories; and
(iii) \textit{Vanilla}, which directly queries the solver without persistent updates.
All evolution methods learn updates on the training set and are evaluated on the test set.
% We use Claude Sonnet 4.5 as the evolver and evaluate solvers ranging from Claude Haiku 4.5 to Claude Sonnet 4/4.5, GPT-5, and Gemini 3 Flash. All other settings follow Sec.~\ref{sec:evaluation_setup}.

\noindent\textbf{Results.}
Tab.~\ref{tab:effectiveness} reports TGC and APT across methods.
We observe two consistent trends.
\emph{(i) Agentic evolution acts as a capability multiplier.}
\method yields substantial gains across all solver backbones.
For example, it achieves $64\%$ and $82\%$ TGC on Claude Haiku 4.5 and Gemini 3 Flash, respectively, compared to $46\%/52\%$ for AWM and $32\%/56\%$ for Vanilla.
This demonstrates that agentic evolution reliably converts deployment evidence into reusable improvements that static prompt or workflow updates fail to capture.
\emph{(ii) Narrowing the capacity gap.}
Smaller solvers augmented with \method can match or exceed larger vanilla models (e.g., Haiku 4.5 with \method reaches $64\%$ TGC versus $42\%$ for vanilla Sonnet~4), indicating that persistent procedural refinement across episodes can rival gains from stronger backbones alone.

\noindent\textbf{Case Studies.}
Qualitative analysis of evolution traces reveals why these gains arise.
Non-agentic baselines such as AWM tend to capture only \emph{surface-level patterns}, retrieving raw trajectories that often include irrelevant context or hallucinated reasoning.
In contrast, \method performs \emph{active diagnosis}: the evolver invokes analysis tools to identify underlying failure causes (e.g., hidden API dependencies).
These insights are synthesized into \emph{persistent artifacts} in $\pi_S$, such as verified tools ($\mathcal{T}$) or refined knowledge schemas ($\mathcal{K}$), rather than stored as raw text.
All updates are gated by the validation registry ($\mathcal{V}$), ensuring that only improvements passing regression checks are committed.
At solve-time, this replaces fragile, probabilistic regeneration with robust, governed capability invocation.
Detailed examples are provided in Appendix~\ref{appendix:case_studies_effectiveness}.

\subsection{Feasibility: Reliability of Agentic Evolution Loop}
\label{sec:evaluation_feasibility}

\noindent\textbf{Setting.}
To isolate the contribution of each component in agentic evolution, we compare \method against four ablated variants:
(i) \emph{A-Evolve/D}, which removes \textbf{diagnosis} and proposes updates directly from raw trajectories;
(ii) \emph{A-Evolve/A}, which retains diagnosis but disables \textbf{analysis tools} for aggregating evidence across episodes;
(iii) \emph{A-Evolve/P}, which removes \textbf{planning} and generates updates in a single step without explicit edit structure; and
(iv) \emph{A-Evolve/V}, which removes \textbf{verification} and commits updates without validation gating.
Claude Sonnet~4.5 is fixed as the evolver, with Claude Haiku~4.5 and Gemini~3 Flash as solvers. Other settings follow Sec.~\ref{sec:evaluation_effectiveness}.

\begin{figure}[t]
    % \small
    \centering
    \begin{subfigure}{\panelwidth}
        \includegraphics[width=0.98\linewidth]{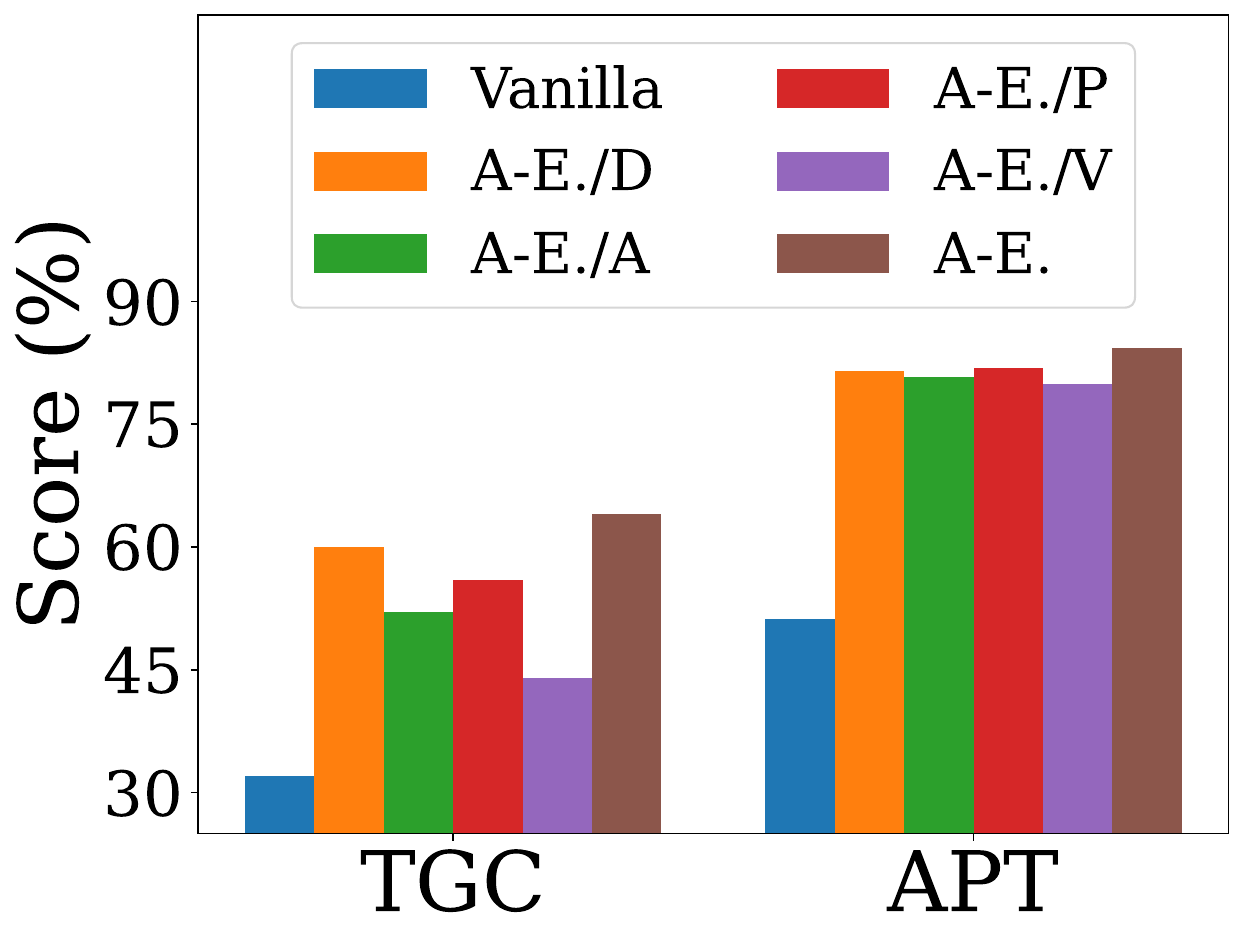}
        \vskip -0.5em
        \caption{Claude Haiku 4.5}
    \end{subfigure}
    \begin{subfigure}{\panelwidth}
        \includegraphics[width=0.98\linewidth]{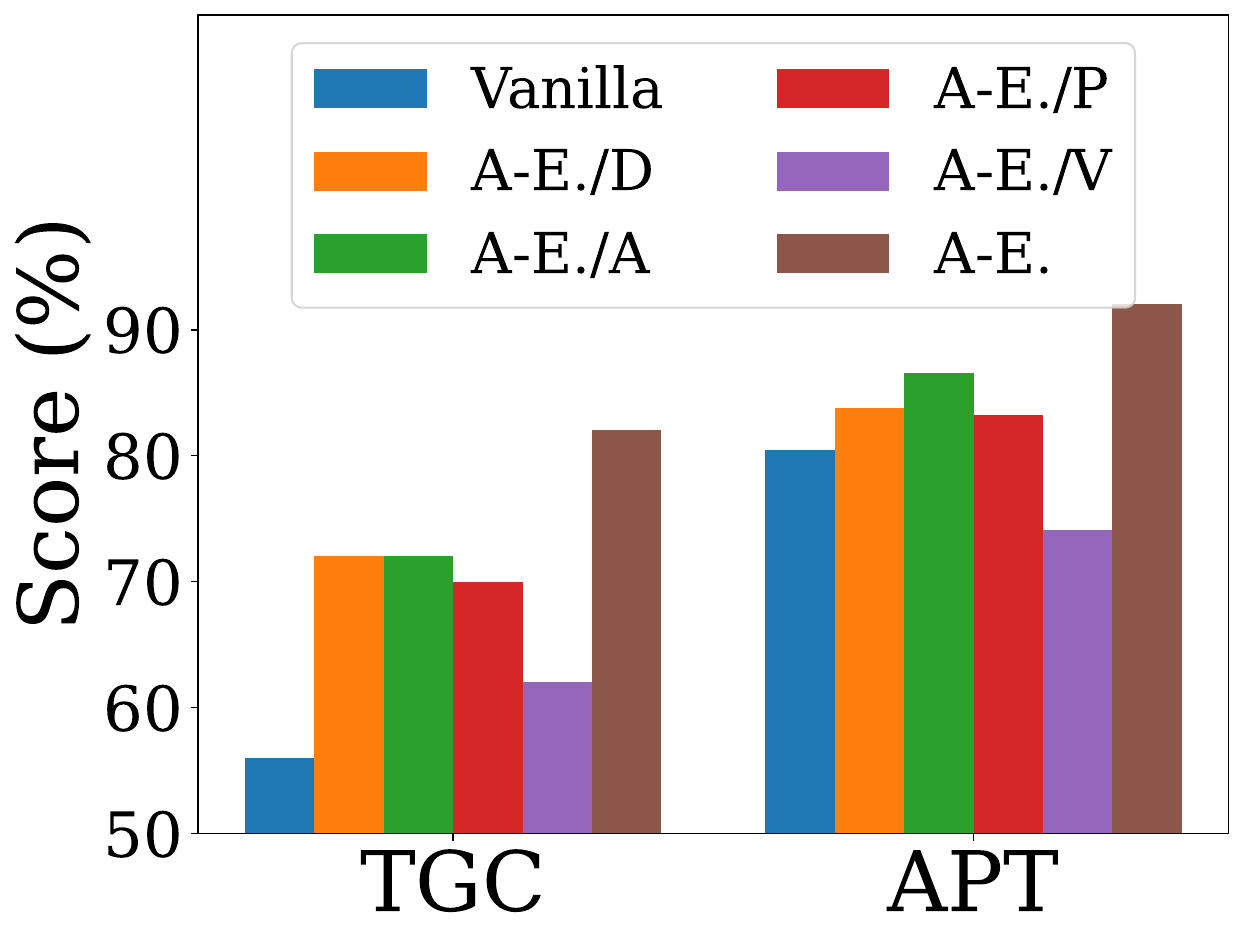}
        \vskip -0.5em
        \caption{Gemini 3 Flash}
    \end{subfigure}
    \vskip -0.5em
    \caption{Ablation studies of \method using Claude Haiku 4.5 and Gemini 3 Flash as solvers.}
    % \vskip -0.5em
    \label{fig:ablation_feasibility}
\end{figure}

\noindent\textbf{Results Analysis.}
Fig.~\ref{fig:ablation_feasibility} shows two consistent trends.
\emph{(i) The full loop is strongest.}
The complete \method achieves the best performance across solvers. Removing any single component degrades results, although all ablations still outperform the vanilla solver, highlighting the complementary roles of the modules and the importance of end-to-end coupling.
\emph{(ii) Verification is a critical stabilizer.}
Among all variants, \emph{A-Evolve/V} degrades the most. Committing updates without validation allows brittle or overfit patches to accumulate, leading to regressions and unstable long-horizon behavior.

\noindent\textbf{Case Studies.}
Qualitative analysis reveals distinct failure modes for each ablation.
Without \emph{diagnosis} (\emph{A-Evolve/D}), the evolver acts blindly, producing superficial patches that mask errors rather than fixing root causes.
Without \emph{analysis tools} (\emph{A-Evolve/A}), updates rely on single-trajectory inference and capture only local symptoms, missing systematic patterns across episodes.
Without \emph{planning} (\emph{A-Evolve/P}), the evolver fails to coordinate interdependent edits, leading to incoherent updates such as modifying tools without updating corresponding schemas.
Finally, removing \emph{verification} (\emph{A-Evolve/V}) results in defective artifacts (e.g., tools with syntax errors) being committed, polluting context and degrading established capabilities.
Detailed examples are provided in Appendix~\ref{appendix:case_studies_feasibility}.

\subsection{Analysis of Evolution Scaling}
\label{sec:meta_scaling_analysis}

% \hanqing{if it is possible, can we change the x-axis from training set size to evolution steps? it's implicitly means the training set size, but focus on the compute not the training data (information)} 
% We vary the training set size from 10 to 120 episodes as a proxy for $C_{\mathrm{evolve}}$, using Claude Haiku~4.5 as the solver and Claude Sonnet~4.5 as the evolver. Other settings follow Sec.~\ref{sec:evaluation_effectiveness}. 

\begin{figure}[t]
    % \small
    \centering
    \begin{subfigure}{\panelwidth}
        \includegraphics[width=0.99\linewidth]{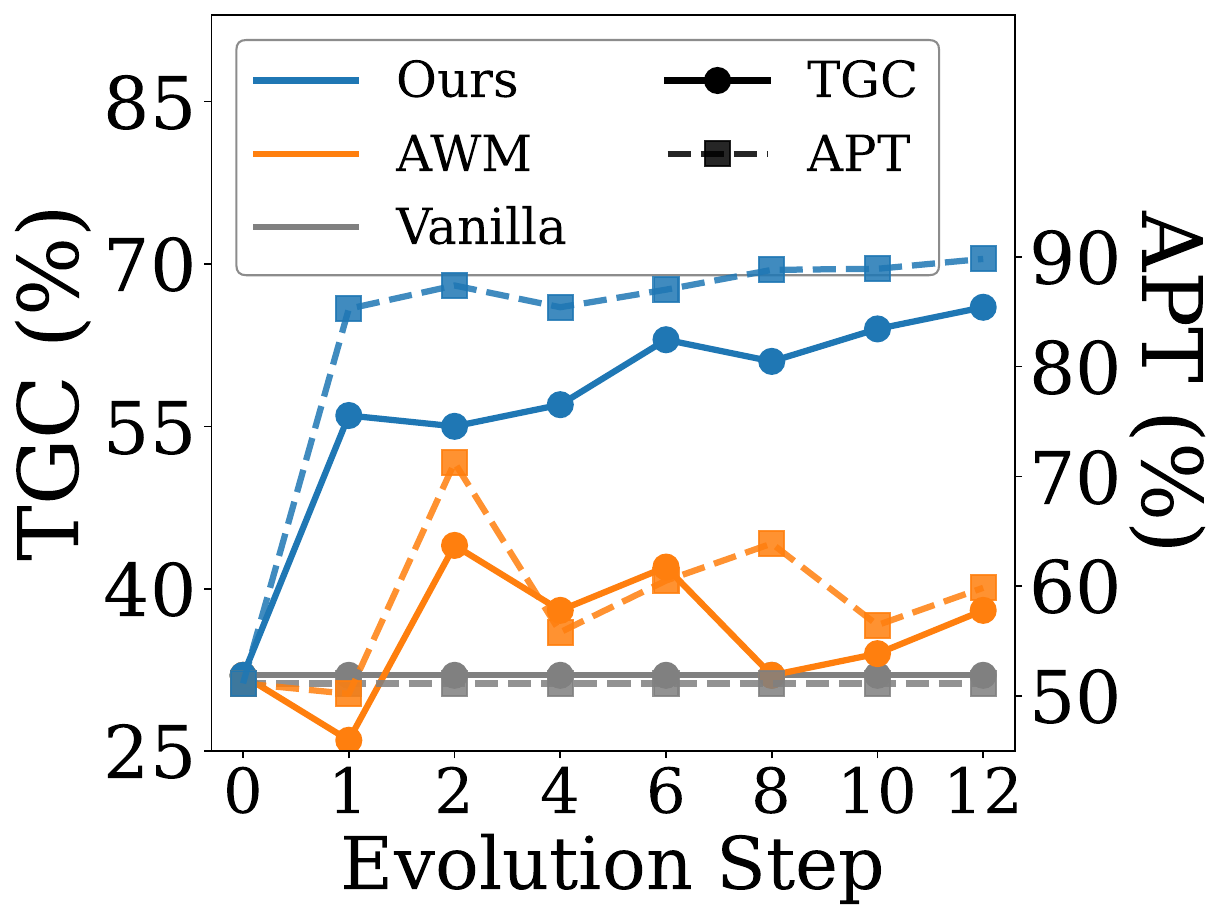}
        \vskip -0.5em
        \caption{Evolution step}
    \end{subfigure}
    \begin{subfigure}{\panelwidth}
        \includegraphics[width=0.99\linewidth]{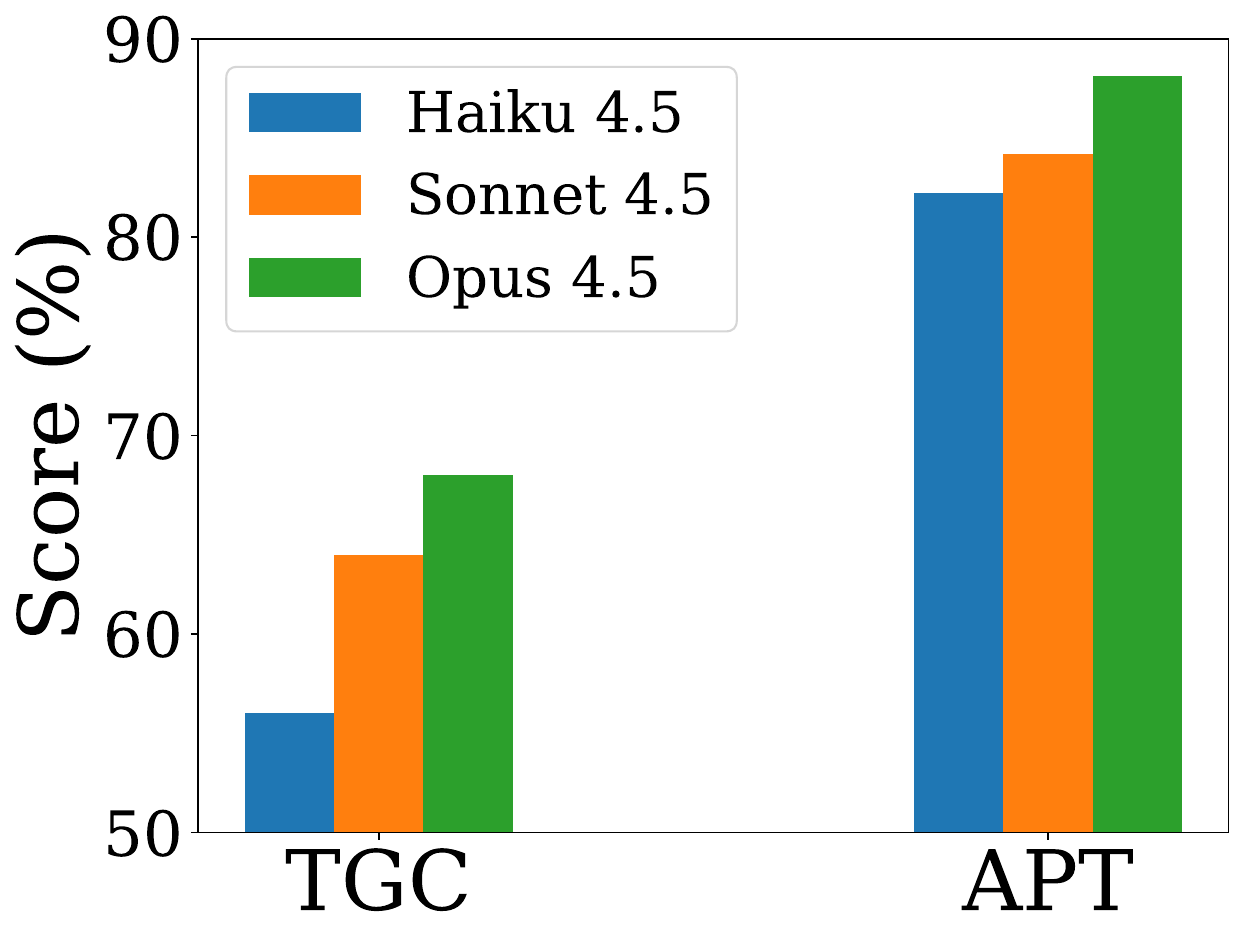}
        \vskip -0.5em
        \caption{Evolver size}
    \end{subfigure}
    % \vskip -0.5em
    \caption{Analysis of evolution-scaling hypotheses: we vary evolution step and evolver model size.}
    \vskip -1.5em
    \label{fig:meta_scaling}
\end{figure}

\noindent\textbf{Scaling Evolution Compute.}
We first evaluate whether increasing evolution-time compute reliably improves capability. We vary the \emph{evolution step} from 1 to 12 steps as a proxy for $C_{\mathrm{evolve}}$, where each step corresponds to the evolver processing a batch of 10 episodes. We use Claude Haiku~4.5 as the solver and Claude Sonnet~4.5 as the evolver. Other settings follow Sec.~\ref{sec:evaluation_effectiveness}. 
Fig.~\ref{fig:meta_scaling}(a) compares \method with vanilla and AWM baselines. Two trends emerge.  
\emph{(i) Sustained scaling.} \method improves monotonically as compute increases, achieving the best performance across all budgets, while AWM quickly plateaus. This supports the evolution-scaling hypothesis that additional evolution compute translates into higher attainable performance, keeping $P^*$ increasing with $C_{\mathrm{evolve}}$.  
\emph{(ii) High efficiency.} Even minimal evolution step (e.g., $0 \!\rightarrow\! 1$ step) yield substantial gains, indicating that structured evolution converts compute into capability more efficiently than heuristic baselines.

\noindent\textbf{Impact of Evolver Size.}
We next examine scaling compute \emph{per update} by increasing \emph{evolver model size}. Fixing the solver as Claude Haiku~4.5, we vary the evolver backbone across Claude Haiku~4.5, Sonnet~4.5, and Opus~4.5; other settings follow Sec.~\ref{sec:evaluation_effectiveness}. Results are shown in Fig.~\ref{fig:meta_scaling}(b).  
\emph{(i) Larger evolvers perform better.} Both TGC and APT increase monotonically with evolver size, mirroring the trends observed under sample scaling and confirming that evolver capacity is a key contributor to $C_{\mathrm{evolve}}$.  
\emph{(ii) Reduced optimization noise.} Qualitative analysis shows that smaller evolvers more often hallucinate root causes and propose brittle updates that fail verification. Larger evolvers more reliably diagnose failures and synthesize robust, generalizable artifacts, converting compute into capability more effectively. Additional analysis is provided in Appendix~\ref{appendix:case_studies_analysis_meta_scaling}.

\vspace{-0.5em}
\section{Alternative Views}

% We now address several alternative perspectives on deployment-time adaptation that frequently arise in discussions of continual learning for LLMs.

We now address three common alternative perspectives on deployment-time adaptation.

\noindent\textbf{``Inference-time reasoning is sufficient.''}
A common counter-argument suggests that explicit evolution is unnecessary if inference-time compute is scaled, allowing models to bridge environmental gaps by ``thinking longer''~\cite{openai2024o1}. While extended reasoning is effective for novel instances, it is inefficient for recurrent failures. Without evolution, the system repeatedly rediscovers the same solutions. Agentic evolution amortizes this cost by converting transient reasoning into persistent artifacts, such as verified tools, yielding stable and reusable capability.

\noindent\textbf{``Why not rely on parametric plasticity?''}
Some advocate direct adaptation through online parameter updates, arguing that continual fine-tuning should replace artifact management~\cite{luo2025empirical}. We contend that unconstrained weight updates are opaque to auditing and prone to catastrophic forgetting. In contrast, agentic evolution operates over explicit, verifiable artifacts (e.g., code and tests), enabling modular updates, structured governance, and stronger safety guarantees than just gradient descent.

\noindent\textbf{``Explicit evolution is too costly.''} Critics argue that maintaining an evolver agent is computationally wasteful compared to heuristics or passive memory. We view this as a trade-off between short-term efficiency and long-term scalability. While heuristics plateau, our results in Sec.~\ref{sec:meta_scaling_analysis} show that agentic evolution continues to improve with evolution-time compute, approaching the compute-optimal frontier. Allocating compute to evolution converts raw compute into durable capability for unbounded environments.

\vspace{-.5em}
\section{Conclusion and Future Directions}

This paper argues that the primary bottleneck to deploying intelligent systems in the real world is not static model capacity, but the ability to \emph{evolve}. As agents move beyond curated training distributions into open-ended environments, static optimization and non-agentic update rules inevitably saturate. We contend that agentic evolution provides a scalable alternative by elevating evolution from a fixed pipeline to an autonomous, governed decision process that converts deployment experience into persistent capability. We further propose the evolution-scaling hypothesis, which posits that adaptation capacity scales with evolution-time compute, introducing a new axis of scaling beyond training-time and inference-time computation.

Realizing the full potential of agentic evolution requires progress along three complementary directions.

\emph{Benchmarks.}
Future benchmarks should explicitly capture the train--deploy gap by placing agents in high-entropy, non-stationary environments where failures cannot be resolved through inference alone. Beyond task success, benchmarks should measure the durability and reuse of evolved artifacts to assess whether evolution-time compute produces lasting capability gains.

\emph{Frameworks.}
Our analysis suggests that the structure of the evolution framework determines how efficiently a system approaches the compute-optimal frontier $P^*$. While \method provides a concrete starting point, future frameworks should further improve diagnosis, planning, and verification to better convert evolution-time compute into stable improvements.

\emph{Theory.}
A foundational theory of agentic evolution remains open. Promising directions include formalizing evolution as optimization over a combinatorial program space and establishing separation results showing that agentic evolution admits a higher attainable frontier than non-agentic heuristics. Bounding regret relative to idealized oracle fine-tuning would provide a principled basis for understanding long-horizon adaptation.

\clearpage
\bibliographystyle{assets/plainnat}
\bibliography{sample-base}

@article{gao2025survey,
  title={A survey of self-evolving agents: On path to artificial super intelligence},
  author={Gao, Huan-ang and Geng, Jiayi and Hua, Wenyue and Hu, Mengkang and Juan, Xinzhe and Liu, Hongzhang and Liu, Shilong and Qiu, Jiahao and Qi, Xuan and Wu, Yiran and others},
  journal={arXiv preprint arXiv:2507.21046},
  year={2025}
}

@article{xia2025agent0,
  title={Agent0: Unleashing self-evolving agents from zero data via tool-integrated reasoning},
  author={Xia, Peng and Zeng, Kaide and Liu, Jiaqi and Qin, Can and Wu, Fang and Zhou, Yiyang and Xiong, Caiming and Yao, Huaxiu},
  journal={arXiv preprint arXiv:2511.16043},
  year={2025}
}

@article{cai2025training,
  title={Training-free group relative policy optimization},
  author={Cai, Yuzheng and Cai, Siqi and Shi, Yuchen and Xu, Zihan and Chen, Lichao and Qin, Yulei and Tan, Xiaoyu and Li, Gang and Li, Zongyi and Lin, Haojia and others},
  journal={arXiv preprint arXiv:2510.08191},
  year={2025}
}

@article{pei2025scope,
  title={SCOPE: Prompt Evolution for Enhancing Agent Effectiveness},
  author={Pei, Zehua and Zhen, Hui-Ling and Kai, Shixiong and Pan, Sinno Jialin and Wang, Yunhe and Yuan, Mingxuan and Yu, Bei},
  journal={arXiv preprint arXiv:2512.15374},
  year={2025}
}

@article{huang2025r,
  title={R-zero: Self-evolving reasoning llm from zero data},
  author={Huang, Chengsong and Yu, Wenhao and Wang, Xiaoyang and Zhang, Hongming and Li, Zongxia and Li, Ruosen and Huang, Jiaxin and Mi, Haitao and Yu, Dong},
  journal={arXiv preprint arXiv:2508.05004},
  year={2025}
}

@article{trivedi2024appworld,
  title={Appworld: A controllable world of apps and people for benchmarking interactive coding agents},
  author={Trivedi, Harsh and Khot, Tushar and Hartmann, Mareike and Manku, Ruskin and Dong, Vinty and Li, Edward and Gupta, Shashank and Sabharwal, Ashish and Balasubramanian, Niranjan},
  journal={arXiv preprint arXiv:2407.18901},
  year={2024}
}

@article{cao2025remember,
  title={Remember me, refine me: A dynamic procedural memory framework for experience-driven agent evolution},
  author={Cao, Zouying and Deng, Jiaji and Yu, Li and Zhou, Weikang and Liu, Zhaoyang and Ding, Bolin and Zhao, Hai},
  journal={arXiv preprint arXiv:2512.10696},
  year={2025}
}

@inproceedings{zhou2022large,
  title={Large language models are human-level prompt engineers},
  author={Zhou, Yongchao and Muresanu, Andrei Ioan and Han, Ziwen and Paster, Keiran and Pitis, Silviu and Chan, Harris and Ba, Jimmy},
  booktitle={The eleventh international conference on learning representations},
  year={2022}
}

@article{wang2024agent,
  title={Agent workflow memory},
  author={Wang, Zora Zhiruo and Mao, Jiayuan and Fried, Daniel and Neubig, Graham},
  journal={arXiv preprint arXiv:2409.07429},
  year={2024}
}

@misc{anthropic2025claudesonnet45,
  author       = {Anthropic},
  title        = {Claude Sonnet 4.5 System Card},
  year         = {2025},
  month        = {Sept},
  url          = {https://assets.anthropic.com/m/12f214efcc2f457a/original/Claude-Sonnet-4-5-System-Card.pdf}
}

@misc{anthropic2025claudehaiku45,
  author       = {Anthropic},
  title        = {Claude Haiku 4.5 System Card},
  year         = {2025},
  month        = {Oct},
  url          = {https://assets.anthropic.com/m/99128ddd009bdcb/Claude-Haiku-4-5-System-Card.pdf}
}

@misc{google2025gemini3flash,
  author       = {Google},
  title        = {Gemini 3 Flash Model Card},
  year         = {2025},
  month        = {Dec},
  url          = {https://storage.googleapis.com/deepmind-media/Model-Cards/Gemini-3-Flash-Model-Card.pdf}
}

@inproceedings{
snell2025scaling,
title={Scaling {LLM} Test-Time Compute Optimally Can be More Effective than Scaling Parameters for Reasoning},
author={Charlie Victor Snell and Jaehoon Lee and Kelvin Xu and Aviral Kumar},
booktitle={The Thirteenth International Conference on Learning Representations},
year={2025},
url={https://openreview.net/forum?id=4FWAwZtd2n}
}

@article{kaplan2020scaling,
  title={Scaling laws for neural language models},
  author={Kaplan, Jared and McCandlish, Sam and Henighan, Tom and Brown, Tom B and Chess, Benjamin and Child, Rewon and Gray, Scott and Radford, Alec and Wu, Jeffrey and Amodei, Dario},
  journal={arXiv preprint arXiv:2001.08361},
  year={2020}
}

@article{singh2025openai,
  title={OpenAI GPT-5 System Card},
  author={Singh, Aaditya and Fry, Adam and Perelman, Adam and Tart, Adam and Ganesh, Adi and El-Kishky, Ahmed and McLaughlin, Aidan and Low, Aiden and Ostrow, AJ and Ananthram, Akhila and others},
  journal={arXiv preprint arXiv:2601.03267},
  year={2025}
}

@article{wei2022chain,
  title={Chain-of-thought prompting elicits reasoning in large language models},
  author={Wei, Jason and Wang, Xuezhi and Schuurmans, Dale and Bosma, Maarten and Xia, Fei and Chi, Ed and Le, Quoc V and Zhou, Denny and others},
  journal={Advances in neural information processing systems},
  volume={35},
  pages={24824--24837},
  year={2022}
}

@article{touvron2023llama,
  title={Llama: Open and efficient foundation language models},
  author={Touvron, Hugo and Lavril, Thibaut and Izacard, Gautier and Martinet, Xavier and Lachaux, Marie-Anne and Lacroix, Timoth{\'e}e and Rozi{\`e}re, Baptiste and Goyal, Naman and Hambro, Eric and Azhar, Faisal and others},
  journal={arXiv preprint arXiv:2302.13971},
  year={2023}
}

@article{radford2018improving,
  title={Improving language understanding by generative pre-training},
  author={Radford, Alec and Narasimhan, Karthik and Salimans, Tim and Sutskever, Ilya and others},
  year={2018},
  publisher={San Francisco, CA, USA}
}

@article{ouyang2025reasoningbank,
  title={Reasoningbank: Scaling agent self-evolving with reasoning memory},
  author={Ouyang, Siru and Yan, Jun and Hsu, I and Chen, Yanfei and Jiang, Ke and Wang, Zifeng and Han, Rujun and Le, Long T and Daruki, Samira and Tang, Xiangru and others},
  journal={arXiv preprint arXiv:2509.25140},
  year={2025}
}

@inproceedings{
schaeffer2025how,
title={How Do Large Language Monkeys Get Their Power (Laws)?},
author={Rylan Schaeffer and Joshua Kazdan and John Hughes and Jordan Juravsky and Sara Price and Aengus Lynch and Erik Jones and Robert Kirk and Azalia Mirhoseini and Sanmi Koyejo},
booktitle={Forty-second International Conference on Machine Learning},
year={2025},
url={https://openreview.net/forum?id=QqVZ28qems}
}

@article{luo2025empirical,
  title={An empirical study of catastrophic forgetting in large language models during continual fine-tuning},
  author={Luo, Yun and Yang, Zhen and Meng, Fandong and Li, Yafu and Zhou, Jie and Zhang, Yue},
  journal={IEEE Transactions on Audio, Speech and Language Processing},
  year={2025},
  publisher={IEEE}
}

@article{shinn2023reflexion,
  title={Reflexion: Language agents with verbal reinforcement learning},
  author={Shinn, Noah and Cassano, Federico and Gopinath, Ashwin and Narasimhan, Karthik and Yao, Shunyu},
  journal={Advances in Neural Information Processing Systems},
  volume={36},
  pages={8634--8652},
  year={2023}
}

@article{openai2024o1,
  title={OpenAI o1 system card},
  author={OpenAI, Aaron Jaech and Kalai, Adam and Lerer, Adam and Richardson, Adam and El-Kishky, Ahmed and Low, Aiden and Helyar, Alec and Madry, Aleksander and Beutel, Alex and Carney, Alex and others},
  journal={arXiv preprint arXiv:2412.16720},
  year={2024}
}

@article{xu2025mem,
  title={A-mem: Agentic memory for llm agents},
  author={Xu, Wujiang and Liang, Zujie and Mei, Kai and Gao, Hang and Tan, Juntao and Zhang, Yongfeng},
  journal={arXiv preprint arXiv:2502.12110},
  year={2025}
}

@article{brown2020language,
  title={Language models are few-shot learners},
  author={Brown, Tom and Mann, Benjamin and Ryder, Nick and Subbiah, Melanie and Kaplan, Jared D and Dhariwal, Prafulla and Neelakantan, Arvind and Shyam, Pranav and Sastry, Girish and Askell, Amanda and others},
  journal={Advances in neural information processing systems},
  volume={33},
  pages={1877--1901},
  year={2020}
}

@inproceedings{
hu2025testtime,
title={Test-Time Learning for Large Language Models},
author={Jinwu Hu and Zitian Zhang and Guohao Chen and Xutao Wen and Chao Shuai and Wei Luo and Bin Xiao and Yuanqing Li and Mingkui Tan},
booktitle={Forty-second International Conference on Machine Learning},
year={2025},
url={https://openreview.net/forum?id=iCYbIaGKSR}
}

@article{gama2014concept,
author = {Gama, Jo\~{a}o and \v{Z}liobaitundefined, Indrundefined and Bifet, Albert and Pechenizkiy, Mykola and Bouchachia, Abdelhamid},
title = {A survey on concept drift adaptation},
year = {2014},
issue_date = {April 2014},
publisher = {Association for Computing Machinery},
address = {New York, NY, USA},
volume = {46},
number = {4},
issn = {0360-0300},
url = {https://doi.org/10.1145/2523813},
doi = {10.1145/2523813},
journal = {ACM Comput. Surv.},
}

@article{jiang2023selfevolve,
  title={Selfevolve: A code evolution framework via large language models},
  author={Jiang, Shuyang and Wang, Yuhao and Wang, Yu},
  journal={arXiv preprint arXiv:2306.02907},
  year={2023}
}

@article{chhikara2025mem0,
  title={Mem0: Building production-ready ai agents with scalable long-term memory},
  author={Chhikara, Prateek and Khant, Dev and Aryan, Saket and Singh, Taranjeet and Yadav, Deshraj},
  journal={arXiv preprint arXiv:2504.19413},
  year={2025}
}

@article{han2024retrieval,
  title={Retrieval-augmented generation with graphs (graphrag)},
  author={Han, Haoyu and Wang, Yu and Shomer, Harry and Guo, Kai and Ding, Jiayuan and Lei, Yongjia and Halappanavar, Mahantesh and Rossi, Ryan A and Mukherjee, Subhabrata and Tang, Xianfeng and others},
  journal={arXiv preprint arXiv:2501.00309},
  year={2024}
}

@article{hu2025test,
  title={Test-Time Learning for Large Language Models},
  author={Hu, Jinwu and Zhang, Zhitian and Chen, Guohao and Wen, Xutao and Shuai, Chao and Luo, Wei and Xiao, Bin and Li, Yuanqing and Tan, Mingkui},
  journal={arXiv preprint arXiv:2505.20633},
  year={2025}
}

@inproceedings{lai2025survey,
    title = "A Survey of Post-Training Scaling in Large Language Models",
    author = "Lai, Hanyu  and
      Liu, Xiao  and
      Gao, Junjie  and
      Cheng, Jiale  and
      Qi, Zehan  and
      Xu, Yifan  and
      Yao, Shuntian  and
      Zhang, Dan  and
      Du, Jinhua  and
      Hou, Zhenyu  and
      Lv, Xin  and
      Huang, Minlie  and
      Dong, Yuxiao  and
      Tang, Jie",
    booktitle = "Proceedings of the 63rd Annual Meeting of the Association for Computational Linguistics (Volume 1: Long Papers)",
    year = "2025",
}

@article{chen2024self,
  title={Self-play fine-tuning converts weak language models to strong language models},
  author={Chen, Zixiang and Deng, Yihe and Yuan, Huizhuo and Ji, Kaixuan and Gu, Quanquan},
  journal={arXiv preprint arXiv:2401.01335},
  year={2024}
}

@article{liu2025spiral,
  title={SPIRAL: Self-Play on Zero-Sum Games Incentivizes Reasoning via Multi-Agent Multi-Turn Reinforcement Learning},
  author={Liu, Bo and Guertler, Leon and Yu, Simon and Liu, Zichen and Qi, Penghui and Balcells, Daniel and Liu, Mickel and Tan, Cheston and Shi, Weiyan and Lin, Min and others},
  journal={arXiv preprint arXiv:2506.24119},
  year={2025}
}

@article{wang2025space,
  title={SPACE: Noise contrastive estimation stabilizes self-play fine-tuning for large language models},
  author={Wang, Yibo and Chen, Qing-Guo and Xu, Zhao and Luo, Weihua and Zhang, Kaifu and Zhang, Lijun},
  journal={arXiv preprint arXiv:2512.07175},
  year={2025}
}

@inproceedings{
hubotter2025efficiently,
title={Efficiently Learning at Test-Time: Active Fine-Tuning of {LLM}s},
author={Jonas H{\"u}botter and Sascha Bongni and Ido Hakimi and Andreas Krause},
booktitle={The Thirteenth International Conference on Learning Representations},
year={2025},
url={https://openreview.net/forum?id=NS1G1Uhny3}
}

@article{wei2025toward,
  title={Toward Training Superintelligent Software Agents through Self-Play SWE-RL},
  author={Wei, Yuxiang and Sun, Zhiqing and McMilin, Emily and Gehring, Jonas and Zhang, David and Synnaeve, Gabriel and Fried, Daniel and Zhang, Lingming and Wang, Sida},
  journal={arXiv preprint arXiv:2512.18552},
  year={2025}
}

@article{behrouz2024titans,
  title={Titans: Learning to memorize at test time},
  author={Behrouz, Ali and Zhong, Peilin and Mirrokni, Vahab},
  journal={arXiv preprint arXiv:2501.00663},
  year={2024}
}

@article{yin2024lofit,
  title={Lofit: Localized fine-tuning on llm representations},
  author={Yin, Fangcong and Ye, Xi and Durrett, Greg},
  journal={Advances in Neural Information Processing Systems},
  volume={37},
  pages={9474--9506},
  year={2024}
}

@article{behrouz2025s,
  title={It's All Connected: A Journey Through Test-Time Memorization, Attentional Bias, Retention, and Online Optimization},
  author={Behrouz, Ali and Razaviyayn, Meisam and Zhong, Peilin and Mirrokni, Vahab},
  journal={arXiv preprint arXiv:2504.13173},
  year={2025}
}

@inproceedings{prasad2023grips,
  title={Grips: Gradient-free, edit-based instruction search for prompting large language models},
  author={Prasad, Archiki and Hase, Peter and Zhou, Xiang and Bansal, Mohit},
  booktitle={Proceedings of the 17th Conference of the European Chapter of the Association for Computational Linguistics},
  pages={3845--3864},
  year={2023}
}

@InProceedings{fernando2024promptbreeder,
  title = 	 {Promptbreeder: Self-Referential Self-Improvement via Prompt Evolution},
  author =       {Fernando, Chrisantha and Banarse, Dylan Sunil and Michalewski, Henryk and Osindero, Simon and Rockt\"{a}schel, Tim},
  booktitle = 	 {Proceedings of the 41st International Conference on Machine Learning},
  pages = 	 {13481--13544},
  year = 	 {2024}
}

@article{yuksekgonul2024textgrad,
  title={Textgrad: Automatic" differentiation" via text},
  author={Yuksekgonul, Mert and Bianchi, Federico and Boen, Joseph and Liu, Sheng and Huang, Zhi and Guestrin, Carlos and Zou, James},
  journal={arXiv preprint arXiv:2406.07496},
  year={2024}
}

@article{shi2025youtu,
  title={Youtu-Agent: Scaling Agent Productivity with Automated Generation and Hybrid Policy Optimization},
  author={Shi, Yuchen and Cai, Yuzheng and Cai, Siqi and Xu, Zihan and Chen, Lichao and Qin, Yulei and Zhou, Zhijian and Fei, Xiang and Qiu, Chaofan and Tan, Xiaoyu and others},
  journal={arXiv preprint arXiv:2512.24615},
  year={2025}
}

@article{zhai2025agentevolver,
  title={Agentevolver: Towards efficient self-evolving agent system},
  author={Zhai, Yunpeng and Tao, Shuchang and Chen, Cheng and Zou, Anni and Chen, Ziqian and Fu, Qingxu and Mai, Shinji and Yu, Li and Deng, Jiaji and Cao, Zouying and others},
  journal={arXiv preprint arXiv:2511.10395},
  year={2025}
}

@article{zhang2025agentic,
  title={Agentic context engineering: Evolving contexts for self-improving language models},
  author={Zhang, Qizheng and Hu, Changran and Upasani, Shubhangi and Ma, Boyuan and Hong, Fenglu and Kamanuru, Vamsidhar and Rainton, Jay and Wu, Chen and Ji, Mengmeng and Li, Hanchen and others},
  journal={arXiv preprint arXiv:2510.04618},
  year={2025}
}

@article{su2025scaling,
  title={Scaling agents via continual pre-training},
  author={Su, Liangcai and Zhang, Zhen and Li, Guangyu and Chen, Zhuo and Wang, Chenxi and Song, Maojia and Wang, Xinyu and Li, Kuan and Wu, Jialong and Chen, Xuanzhong and others},
  journal={arXiv preprint arXiv:2509.13310},
  year={2025}
}

@inproceedings{wang2025evoagentx,
  title={Evoagentx: An automated framework for evolving agentic workflows},
  author={Wang, Yingxu and Liu, Siwei and Fang, Jinyuan and Meng, Zaiqiao},
  booktitle={Proceedings of the 2025 Conference on Empirical Methods in Natural Language Processing: System Demonstrations},
  pages={643--655},
  year={2025}
}

@article{wei2025evo,
  title={Evo-Memory: Benchmarking LLM Agent Test-time Learning with Self-Evolving Memory},
  author={Wei, Tianxin and Sachdeva, Noveen and Coleman, Benjamin and He, Zhankui and Bei, Yuanchen and Ning, Xuying and Ai, Mengting and Li, Yunzhe and He, Jingrui and Chi, Ed H and others},
  journal={arXiv preprint arXiv:2511.20857},
  year={2025}
}

@Article{brohi2025research,
AUTHOR = {Brohi, Sarfraz and Mastoi, Qurat-ul-ain and Jhanjhi, N. Z. and Pillai, Thulasyammal Ramiah},
TITLE = {A Research Landscape of Agentic AI and Large Language Models: Applications, Challenges and Future Directions},
JOURNAL = {Algorithms},
VOLUME = {18},
YEAR = {2025},
DOI = {10.3390/a18080499}
}

@article{wei2026agentic,
  title={Agentic Reasoning for Large Language Models},
  author={Wei, Tianxin and Li, Ting-Wei and Liu, Zhining and Ning, Xuying and Yang, Ze and Zou, Jiaru and Zeng, Zhichen and Qiu, Ruizhong and Lin, Xiao and Fu, Dongqi and others},
  journal={arXiv preprint arXiv:2601.12538},
  year={2026}
}

@inproceedings{lin2026harness,
    title = "Harness Updating Is Not Harness Benefit: Disentangling Evolution Capabilities in Self-Evolving {LLM} Agents",
    author = "Lin, Minhua  and
      Wu, Juncheng  and
      Wang, Zijun  and
      Shi, Zhan  and
      Sang, Yisi  and
      He, Bing  and
      Liu, Zewen  and
      Wei, Tianxin  and
      Wu, Zongyu  and
      Zhang, Zhiwei  and
      Wang, Dakuo  and
      Zhang, Xiang  and
      Dumoulin, Benoit  and
      Xie, Cihang  and
      Zhou, Yuyin  and
      Wang, Suhang  and
      Lu, Hanqing",
    booktitle = "Proceedings of the 2026 Conference on Empirical Methods in Natural Language Processing",
    year = "2026",
    pages = "52829--52852",
}

@article{lin2026memma,
  title={Memma: Coordinating the memory cycle through multi-agent reasoning and in-situ self-evolution},
  author={Lin, Minhua and Zhang, Zhiwei and Lu, Hanqing and Liu, Hui and Tang, Xianfeng and He, Qi and Zhang, Xiang and Wang, Suhang},
  journal={arXiv preprint arXiv:2603.18718},
  year={2026}
}

@inproceedings{wei2026evo,
  title={Evo-harness: Context-to-harness skill compilation for self-evolving agents},
  author={Wei, Tianxin and Shi, Zhan and Lin, Minhua and He, Bing and Liu, Zewen and Sang, Yisi and Bei, Yuanchen and Ning, Xuying and Zou, Jiaru and Li, Ting-Wei and others},
  booktitle={Proceedings of the 2026 Conference on Empirical Methods in Natural Language Processing},
  pages={22343--22358},
  year={2026}
}

\clearpage
\beginappendix
\section{Related Work: Evolution Paradigms of LLMs}
\label{appendix:full_details_evolution_category}
In this section, we review prior work through the lens of our evolution definition in Eq.~\ref{eq:evolution_update_rewrite} and the three orthogonal axes: \emph{what} is updated (parametric weights $\pi_\theta$ vs.\ persistent artifact state $\pi_S$), \emph{when} updates occur (offline training vs.\ deployment-time), and \emph{how} the update rule $F_{\mathrm{Evolve}}$ is realized (fixed heuristics vs.\ an explicit agentic optimizer). The evolution paradigm is outlined in Tab.~\ref{tab:evolution-comparison}.

\begin{table*}[h]
\centering
% \scriptsize
\small
% \vspace{-0.6em}
\caption{Evolution paradigms organized by \emph{what} is updated, \emph{when} updates occur, and \emph{how} the update rule $F_{\mathrm{Evolve}}$ is instantiated.}
% \vspace{-0.4em}
\label{tab:evolution-comparison}
% {\begin{tabularx}{0.80\textwidth}{ p{3.2cm} | p{1.8cm} | p{2.55cm} | X }
{\begin{tabularx}{0.83\textwidth}{ p{4cm} | p{1.8cm} | p{2.55cm} | X }
\toprule
\textbf{Paradigm} &
\textbf{What is Updated} &
\textbf{When is Updated} &
\textbf{How $F_{\mathrm{Evolve}}$ is Realized} \\
\midrule

\textbf{Offline Parametric Evolution} &
$\pi_\theta$ &
Offline (pre-deployment) &
Training loop (e.g., self-play / distillation) \\
\midrule

\textbf{Online Parametric Evolution} &
$\pi_\theta$  &
Online (deployment-time) &
Online optimization (e.g., gradient descent) \\
\midrule

\textbf{Non-parametric Heuristic Evolution} &
$\pi_S$ ($\mathcal{K}$) &
Online (deployment-time) &
% Fixed heuristic rules (e.g., append-retrieve, prompt search) \\
Fixed heuristic rules \\
\midrule

\textbf{Agentic Evolution (Ours)} &
\textbf{$\pi_S$ ($\mathcal{K}, \mathcal{T}, \mathcal{V}$) } &
Online (deployment-time) &
\textbf{Explicit evolver agent} \\
\bottomrule
\end{tabularx}}
\vspace{-0.6em}
\end{table*}

\subsection{Offline Parametric Evolution}
\paragraph{Self-play fine-tuning from synthetic interactions.}
A predominant paradigm for enhancing agent capability is to improve the parametric backbone $\pi_\theta$ \emph{offline} by bootstrapping training signals from the model itself.
Traditional approaches, such as SPIN~\cite{chen2024self}, iteratively refine the policy by contrasting self-generated outputs against a target distribution.
Recent advances extend this to multi-turn and multi-agent settings: SPIRAL~\cite{liu2025spiral} induces reasoning skills via self-play in structured interaction games, while SPACE~\cite{wang2025space} stabilizes the self-play objective to mitigate training instability.
While these methods enlarge pre-deployment capability, they commit improvements solely to $\pi_\theta$ during training, leaving the agent static and unable to adapt to distribution shifts once deployed.

\noindent\textbf{Zero-data curriculum generation.}
To circumvent the reliance on human-curated datasets, recent frameworks construct self-generated curricula within the offline loop.
R-Zero~\cite{huang2025r} and Agent0~\cite{xia2025agent0} instantiate co-evolving roles (e.g., a challenger and a solver) to synthesize frontier tasks and tool-use trajectories without external data.
Similarly, in software engineering domains, SSR~\cite{wei2025toward} trains agents by injecting and repairing bugs in sandboxed repositories.
Although these methods effectively address the data bottleneck, they remain instances of \emph{offline parametric evolution}: adaptation ceases the moment the model is deployed.

\noindent\textbf{Agentic Continual Pre-training.}
Beyond fine-tuning, recent works propose injecting agentic capabilities~\cite{brohi2025research,wei2026agentic} even earlier in the pipeline.
AgentFounder~\cite{su2025scaling} introduces \emph{Agentic Continual Pre-training (CPT)} as an intermediate stage, scaling agentic capabilities via massive offline data synthesis (e.g., First-order and Higher-order Action Synthesis) derived from knowledge bases and trajectories.
While this approach significantly raises the baseline performance of the foundation model, it remains fundamentally an instance of \emph{offline parametric evolution}: the model's adaptability is frozen once the pre-training concludes, leaving it vulnerable to novel environmental shifts during deployment.

\subsection{Online Parametric Evolution}
\noindent\textbf{Test-time fine-tuning.}
Online parametric evolution attempts to update $\pi_\theta$ (or a subspace thereof) during deployment.
Approaches like SIFT~\cite{hubotter2025efficiently} explore active fine-tuning by selecting informative test-time examples for weight updates.
To reduce computational overhead, methods such as LoFiT~\cite{yin2024lofit} localize updates to specific internal representations or heads.
While enabling continuous adaptation, direct parametric modification introduces significant governance challenges, such as catastrophic forgetting~\cite{luo2025empirical}, irreversibility, and instability, making it risky for long-horizon deployment.

\noindent\textbf{Architectures for test-time memorization.}
A complementary direction blurs the boundary between weights and state by updating neural memory modules during streaming inference.
Titans~\cite{behrouz2024titans} introduces a neural long-term memory module for context memorization, while MIRAS~\cite{behrouz2025s} provides a framework for optimizing associative memory online.
In our taxonomy, these methods still rely on \emph{parametric} updates to a learnable module rather than explicit updates to interpretable artifacts, retaining the opacity of weight-based evolution.

\subsection{Heuristic Non-Parametric Evolution}
\noindent\textbf{Experience memory with fixed retrieval.}
To avoid the risks of weight updates, many systems~\cite{ouyang2025reasoningbank,wei2025evo,xu2025mem} modify a persistent artifact state $\pi_S$, typically a memory bank of traces or summaries, using fixed heuristics.
ReasoningBank~\cite{ouyang2025reasoningbank} distills strategies into a reusable memory bank via self-judgment.
% , while COPER~\cite{coper2025} couples prompt evolution with an experience replay buffer.
However, because the update rule $F_{\mathrm{Evolve}}$ is heuristic (e.g., ``append and retrieve''), these systems often suffer from context saturation and retrieval noise as experience accumulates.

\noindent\textbf{Prompt and context optimization.}
Another class of non-parametric methods~\cite{prasad2023grips, fernando2024promptbreeder,zhou2022large} treats $\pi_S$ as a set of optimizable prompts.
GrIPS~\cite{prasad2023grips} and Promptbreeder~\cite{fernando2024promptbreeder} apply discrete search operators (e.g., mutation, crossover) to evolve prompts.
More recently, Training-Free GRPO~\cite{cai2025training} performs policy optimization in the context space via token priors, and TextGrad~\cite{yuksekgonul2024textgrad} applies textual differentiation to optimize components.
While effective, these methods typically employ a static, pre-specified search algorithm as $F_{\mathrm{Evolve}}$, limiting the system's ability to strategically diagnose failures or plan complex structural updates.

\subsection{Toward Agentic Evolution}

\noindent\textbf{Agent-driven context and prompt engineering.}
Several works~\cite{lin2026harness,wei2026evo} operationalize prompt engineering as an agentic reasoning task.
ACE~\cite{zhang2025agentic} and SCOPE~\cite{pei2025scope} treat context management as an online optimization problem, employing a modular loop (generate--reflect--curate) to evolve playbooks or guidelines.
While these methods move beyond heuristic search by using an LLM to reason about updates, they are primarily restricted to the \emph{textual context} modality, optimizing instructions rather than executable code or structural system logic.

\noindent\textbf{Automated synthesis of tools and memory.}
Other approaches~\cite{shi2025youtu,zhai2025agentevolver,wang2025evoagentx,lin2026memma} focus on synthesizing specific functional components.
Youtu-Agent~\cite{shi2025youtu} and AgentEvolver~\cite{zhai2025agentevolver} introduce meta-agents capable of generating new tools or training curricula to expand agent capabilities.
Similarly, A-Mem~\cite{xu2025mem} targets the memory modality, employing an agentic controller to dynamically restructure and cross-reference memory artifacts rather than passively appending logs.
These systems demonstrate the feasibility of agent-driven updates for specific artifacts.
However, they typically operate as specialized pipelines for a single type of state (e.g., only tools or only memory) and often lack a unified mechanism for \emph{governed diagnosis} and \emph{verification} across diverse failure modes.

\noindent\textbf{Positioning of our work.}
Existing works can be viewed as specific instances of using LLMs to optimize state.
Our work generalizes these developments into the formal framework of \textbf{Agentic Evolution}.
Unlike prior approaches that focus on optimizing a single modality (e.g., prompts \emph{or} tools), we propose a holistic, budget-aware evolver capable of diagnosing diverse execution failures and proposing structural edits to a heterogeneous state $\pi_S$.
Crucially, we introduce the principles of \emph{goal orientation}, \emph{autonomy}, and \emph{compositionality} to strictly govern this process, ensuring that agent-driven updates are not just flexible, but reliable and scalable enough for deployment.
% \paragraph{Agentic context engineering.}
% The emerging paradigm of agentic evolution treats the update process itself as a reasoning task performed by an optimizer agent.
% ACE (Agentic Context Engineering)~\cite{ace2025} and SCOPE~\cite{scope2025} formalize context management as an online optimization problem, using a modular loop (generate--reflect--curate) to evolve playbooks and prevent context collapse.
% Unlike heuristic search, the ``optimizer'' here is an LLM capable of semantic reasoning about \emph{how} to improve $\pi_S$.

% \paragraph{Automated agent synthesis and memory.}
% Current research is scaling this principle to evolve the entire agent configuration.
% Youtu-Agent~\cite{youtuagent2025} and AgentEvolver~\cite{agentevolver2025} propose frameworks where a meta-agent continuously synthesizes tools, workflows, and training tasks.
% Similarly, A-Mem~\cite{amem2025} applies agentic governance to long-term memory artifacts.
% Our work unifies these developments under the framework of \emph{Agentic Evolution}, emphasizing that $F_{\mathrm{Evolve}}$ should be designed as a governed, budget-aware, and goal-directed optimizer that transforms raw experience into durable, structured capabilities.

\section{Additional Details of Experimental Setup}
\subsection{Dataset Details}
\label{appenidx:dataset_details}
In this paper, we focus on AppWorld, a controllable environment and benchmark designed for interactive coding agents. It simulates a world of 9 day-to-day applications (e.g., Amazon, Spotify, Venmo, Gmail) and 2 helper apps (ApiDocs, Supervisor) through $457$ APIs. The environment is populated with the digital activities of approximately $100$ fictitious users, simulating realistic lives and relationships (e.g., family, roommates) to support complex interaction scenarios.

The AppWorld framework consists of two primary components: 
\begin{itemize}[leftmargin=*]
\item \textbf{AppWorld engine:} This is a high-quality execution environment. It provides a fully controllable and reproducible simulator where agents can operate apps via APIs without real-world consequences, such as spending actual money or spamming emails. The engine allows agents to execute rich Python code, including loops and conditionals, to interact with the environment iteratively. 
\item \textbf{AppWorld benchmark:} A suite of $750$ diverse and challenging tasks derived from $250$ task scenarios. These tasks require agents to navigate multiple apps (average $1.8$ apps per task) and utilize up to $26$ APIs (avg. $9.5$) with dependencies
between calls (i.e., outputs of calls used as inputs
to subsequent calls). The dataset is split into ``normal'' (\texttt{test-normal}) and ``challenge'' (\texttt{test-challenge}) sets, where the challenge set requires agents to utilize APIs from apps seen only during test time. 
\end{itemize}

\subsection{Implementation Details of Agentic Evolution}
\label{appendix:a_evolve_implementation}
In this subsection, we provide granular details on the architecture, artifact representation, and inference configuration of A-Evolve.

\noindent\textbf{Architecture of the evolver agent.}
The Evolver is implemented as a sequential pipeline with four specialized components: Observer, Proposer, Updater, and Verifier. Unlike role-based prompting, each component is instantiated as a distinct module with dedicated logic and persistence.
\begin{itemize}
    \item \textbf{Diagnoser:} Receives the full interaction trace $\mathrm{Obs}_{1:t}$ from the Solver, including the user instruction, tool calls, stdout/stderr, and the final error. Its objective is \emph{failure attribution}: distinguishing between stochastic environment noise and epistemic gaps in the agent's policy. 
    We employ a cross-episode aggregation strategy over batched observations (window size $W=10$) to perform root cause analysis, filtering out transient errors to focus on persistent capability deficits.
    \item \textbf{Planner:} Conditioned on the diagnosis, the planner is responsible for synthesizing a structural update $\delta$. We constrain the planner's action space to a strict schema of operations $\mathcal{A} = \{ \textsc{CreateTool},\allowbreak \textsc{EvolveTool},\allowbreak \textsc{AddKnowledge},\allowbreak \textsc{EvolveKnowledge},\allowbreak \ldots \}$. This structured output forces the model to map abstract failure modes into concrete, semantically meaningful artifact definitions rather than unstructured text, significantly reducing the search space for improvements.
    \item \textbf{Updater:} The updater acts as the deterministic execution engine. It holds write access to the \emph{Artifact Registry} (a version-controlled file system) and translates the Planner's high-level intent into atomic file operations. It enforces strict separation of concerns, ensuring that tools (Python modules), skills (procedural knowledge), and facts (JSON knowledge base) are stored in their respective canonical formats.
    \item \textbf{Verifier:} Responsible for governance. In our case, it validates syntactic correctness of generated code (Python AST parsing for tools), schema compliance (YAML frontmatter for skills), and runtime safety (isolated execution in \texttt{ToolWorkspace}). Only proposals passing validation are committed to the workspace and reflected in the agent's system prompt via on-policy state injection.
\end{itemize}

\noindent\textbf{Hyperparameters and Model Configuration.}
Unless otherwise specified, we use Claude Sonnet 4.5~\cite{anthropic2025claudesonnet45} as the evolver backbone, and evaluate solvers ranging from Claude Haiku 4.5~\cite{anthropic2025claudehaiku45} and Claude Sonnet 4/4.5 to GPT-5~\cite{singh2025openai} and Gemini 3 Flash~\cite{google2025gemini3flash}. 
For generation hyperparameters, we use a temperature of $0.7$ for both the evolver and solver. The maximum output token limits are set to $8,192$ for the evolver and $4,096$ for the solver.

\subsection{Evaluation Metrics}
\label{appendix:evaluation_metrics}
To ensure the robustness of our evaluation, following prior work~\cite{trivedi2024appworld,cao2025remember}, we report two complementary metrics based on the concept of a \textbf{task score}. 

For each task $i$, we define its score $s_i \in [0, 1]$ as the fraction of passed unit tests:
\begin{equation}
    s_i = \frac{\text{passed\_tests}_i}{\text{total\_tests}_i}.
\end{equation}

Based on this score, we define:

\begin{itemize}[leftmargin=*]
    \item \textbf{Average Passed Tests (APT):} This metric captures the granular, incremental progress of the agent. It is simply the average score across all evaluation tasks:
    \begin{equation}
        \text{APT} = \frac{1}{N} \sum_{i=1}^{N} s_i.
    \end{equation}
    APT is particularly valuable for measuring evolution, as it reflects improvements in capability (e.g., $s_i$ increasing from 0.2 to 0.8) even when the full task is not yet perfectly solved.

    \item \textbf{Task Goal Completion (TGC):} This metric represents the strict success rate. A task is considered successfully completed if and only if it achieves a perfect score (i.e., all verification conditions are met). TGC is defined as:
    \begin{equation}
        \text{TGC} = \frac{1}{N} \sum_{i=1}^{N} \mathbb{I}(s_i = 1),
    \end{equation}
    where $N$ is the total number of tasks and $\mathbb{I}(\cdot)$ is the indicator function.
\end{itemize}

% \emph{Task Goal Completion} (TGC), the fraction of tasks successfully completed, and \emph{Average Passed Tests} (APT), the average fraction of unit tests passed per task. To ensure fair comparison, we fix a solve-time compute budget $C_{\mathrm{solve}}$ (maximum tool calls, steps, or tokens per task) and an evolve-time compute budget $C_{\mathrm{evolve}}$ (maximum tokens and tool invocations per episode) across all methods. Full evaluation details are given in Appendix~\ref{appendix:evaluation_metrics}.

\section{More details of Experimental Results}
\subsection{Case Studies of Effectiveness of Agentic Evolution in Sec.~\ref{sec:evaluation_effectiveness}}
\label{appendix:case_studies_effectiveness}
% \noindent\textbf{Full Case Studies.}
% \subsection{Case Studies: Agentic vs.\ Non-Agentic Evolution}
% \label{appendix:case_studies_effectiveness}
To provide qualitative evidence for the effectiveness trends in Tab.~\ref{tab:effectiveness}, we analyze representative failure and success patterns from A-Evolve and AWM. Both methods use the same training and test sets (50 tasks each) with Claude Haiku~4.5 as the solver.

\noindent\textbf{Observation 1: failure recovery mechanisms.}
A key distinction is how each method handles errors mid-trajectory:

\begin{itemize}[leftmargin=*]
    \item \textbf{AWM (Passive):} When the solver encounters an API error (e.g., \texttt{401 Unauthorized}), it has no mechanism to update its workflow in-place. The error propagates, and the solver must re-explore from scratch---often repeating the same mistake. In task \emph{``What is the title of the most-liked song in my Spotify playlists''}, the agent cycles through 30 identical failed attempts without adaptation.
    
    \item \textbf{A-Evolve (Active):} After observing repeated \texttt{401} errors across a batch, the evolver creates a \texttt{manage\_auth\_token} tool that automatically caches and injects access tokens. The diagnosis explicitly identifies: ``\textit{The trajectory shows repeated 401 errors because the agent forgets to pass `access\_token' after login.}'' This structural insight is encoded as a verified tool, enabling one-shot authentication in subsequent tasks.
\end{itemize}

\vspace{0.5em}
\noindent\textbf{Observation 2: context saturation.}
We compare the learned artifacts after 50 training tasks:
\begin{itemize}[leftmargin=*]
    \item AWM produces a single monolithic workflow file that grows linearly with training. This file contains concatenated trajectory fragments with task-specific details (e.g., hardcoded credentials, specific note IDs) that do not generalize. When retrieved, these fragments often mislead the solver with irrelevant context.
    \item A-Evolve produces 16 modular skills (e.g., \texttt{systematic\_api\_exploration}, \texttt{detect\_execution\_stall}), 4 verified tools (e.g., \texttt{authenticate\_app}), and 4 knowledge entries (e.g., \texttt{appworld\_authentication}). Each artifact is \emph{semantically indexed} by trigger conditions and \emph{verified} before commitment.
\end{itemize}

\noindent\textbf{Observation 3: trajectory comparison.}
For task \emph{``Reply to Christopher with movie recommendations from SimpleNote''}, we compare solve-time behavior:

\emph{AWM Agent (Score: 0.88, 29 steps):}
\begin{itemize}[leftmargin=*]
    \item Steps 1-6: Explores API documentation (redundant with workflow)
    \item Steps 7-11: Fails 401 on SimpleNote, re-explores
    \item Steps 12-23: Successful login and note retrieval after trial-and-error
    \item Steps 24-28: Finds contact, sends message
    \item Step 29: Completes task
\end{itemize}

\emph{A-Evolved Agent (Score: 1.0, 8 steps):}
\begin{itemize}[leftmargin=*]
    \item Step 1: Invokes \texttt{analyze\_task\_requirements()} to parse goal
    \item Step 2: Retrieves \texttt{authentication\_tool\_usage\_pattern} skill
    \item Steps 3-4: Authenticates SimpleNote and Phone with cached token procedure
    \item Steps 5-6: Retrieves note, extracts movies
    \item Step 7: Sends message to Christopher
    \item Step 8: Completes task
\end{itemize}

The A-Evolved agent is $3.6\times$ more efficient (8 vs.\ 29 steps) because it \emph{amortizes} authentication and goal-parsing into reusable procedures that execute deterministically, rather than re-exploring on each task.

\noindent\textbf{Takeaway.}
The case studies reveal a fundamental asymmetry: AWM captures \emph{what} the agent did (raw trajectories), while A-Evolve captures \emph{why} failures occurred and \emph{how} to prevent them (verified tools, diagnostic knowledge). This explains the consistent TGC and APT gaps in Tab.~\ref{tab:effectiveness}: agentic evolution converts deployment feedback into persistent, governed capabilities that static workflow retrieval cannot match.

\subsection{Analysis of Feasibility of Agentic Evolution in Sec.~\ref{sec:evaluation_feasibility}}
% \noindent\textbf{Case Studies}
\label{appendix:case_studies_feasibility}
\noindent\textbf{Qualitative analysis.}
We analyze evolution traces and observe distinct failure modes for each ablation. 
\begin{itemize}[leftmargin=*]
    \item \emph{Diagnosis} is critical for depth: without it (\emph{A-Evolve/D}), the agent acts \textbf{blindly}, forced to infer updates directly from \textbf{raw trajectories} without root-cause reasoning. Consequently, it resorts to \textbf{superficial patching} (e.g., masking a crash with `try-except` rather than fixing the incorrect parser logic).
    \item \emph{Analysis Tools} enable \textbf{pattern discovery}: without them (\emph{A-Evolve/A}), the agent relies on pure LLM inference over single trajectories, capturing only \textbf{superficial symptoms} (e.g., fixing a one-off error message). In contrast, analysis tools allow the evolver to aggregate \textbf{underlying statistical patterns} across episodes (e.g., clustering recurring failure modes), thereby distinguishing systematic deficits from transient noise.
    \item \emph{Planning} ensures coherence: without it (\emph{A-Evolve/P}), the agent lacks the mechanism to manage inter-dependencies between artifacts, leading to disjointed updates (e.g., modifying a tool's signature in $\mathcal{T}$ without updating its usage schema in $\mathcal{K}$, causing subsequent execution failures).
    \item Finally, \emph{Verification} acts as the stabilizer: without it (\emph{A-Evolve/V}), the system blindly commits \textbf{defective artifacts} (e.g., tools with \textbf{syntax errors}). This \textbf{pollutes the context}, causing the solver to waste inference budget on broken calls, and introduces regressions that degrade established capabilities.
\end{itemize}

\noindent\textbf{Case Studies.} We provide several representative case studies to better support our analysis:

\emph{Case 1: verification prevents defective artifacts (A-Evolve/V).}
When considering task \emph{``What is the title of the most-played song in my Spotify album library''}, the evolver created a \texttt{discover\_apis} tool without verification.
On the solver's first invocation:
\begin{quote}
\small
        \textit{discover\_apis: {'success': False, 'error': "module 'appworld' 
    has no attribute 'list\_apis'"}}
\end{quote}
The solver wasted a step before falling back to manual exploration.
This pattern repeated: \textbf{2 broken tools committed} across evolution, causing runtime errors that degraded solver efficiency by $\sim$15\% compared to full A-Evolve.

\emph{Case 2: planning enables implementable fixes (A-Evolve/P).} 
The planning stage translates high-level diagnostic insights into \emph{coordinated action sequences}. 
Consider a task in full A-Evolve: the diagnosis identifies ``agent wastes steps on 401 authentication errors.'' 
The \textbf{Plan} generates 4 interdependent actions:
\begin{itemize}[leftmargin=*]
    \item \texttt{CREATE\_TOOL}: \texttt{discover\_api\_spec} (API discovery)
    \item \texttt{CREATE\_TOOL}: \texttt{manage\_auth\_token} (token storage)
    \item \texttt{ADD\_SKILL}: \texttt{systematic\_api\_exploration} (usage pattern)
    \item \texttt{ADD\_SKILL}: \texttt{authentication\_workflow} (login procedure)
    % \item \texttt{ADD\_KNOWLEDGE}: \texttt{common\_api\_conventions} (naming patterns)
    % \item \texttt{ADD\_KNOWLEDGE}: \texttt{api\_error\_recovery} (recovery strategies)
\end{itemize}
Crucially, item (3) relies on item (1): the skill \texttt{systematic\_api\_exploration} explicitly instructs the solver to invoke the newly created \texttt{discover\_api\_spec} tool.
\textbf{Without planning, this dependency graph is lost.} In the A-Evolve/P ablation, although the diagnosis produced similar insights, the immediate generation step failed to sequence the creation process. It attempted to define the skill before the tool existed or hallucinated a tool signature, resulting in validation failures.
Consequently, while A-Evolve/P accumulated some loose textual knowledge, it successfully synthesized {0 executable tools and 0 structured skills}, failing to implement coherent, multi-artifact solutions.

\emph{Case 3: diagnosis enables root-cause analysis (A-Evolve/D).}
Without diagnosis, the evolver observed the same errors but lacked structured analysis.
With full A-Evolve, the diagnosis identified:
\begin{quote}
\small\textit{``Agent uses wrong parameter names (e.g., `email' instead of `username') causing 422 errors. The agent doesn't consistently check API documentation before calling APIs.''}
\end{quote}
This insight led to the \texttt{systematic\_api\_exploration} skill, reducing 422 errors from 8 per trajectory to $<$1.

\emph{Case 4: analysis tools enable pattern discovery (A-Evolve/A).}
Without analysis tools, the evolver relied on LLM inference over single trajectories.
With analysis tools, \texttt{grep\_observations} revealed that 401 errors occurred in \textbf{18/20 training tasks}, enabling the evolver to prioritize \texttt{authentication\_workflow} over superficial one-off fixes.

\noindent\textbf{Takeaway.}
Reliable agentic evolution is an \emph{irreducible loop}: diagnosis and analysis enable targeted updates, planning enables implementable multi-step fixes, and verification contributes most for stable accumulation over long horizons.

% \section{More Details of Case Studies}
% \subsection{Case Studies of Effectiveness}

% \subsection{Analysis of Evolution-scaling Hypotheses in Sec.~\ref{sec:meta_scaling_analysis}}
\subsection{Case Studies of Impact of Evolver Size}
\label{appendix:case_studies_analysis_meta_scaling}

% \noindent\textbf{Case Studies of Impact of Evolver Size.}
To provide qualitative evidence for the scaling trends in Fig.~\ref{fig:meta_scaling}(b), we analyze how evolvers of different capacities respond to identical failure patterns. We run A-Evolve on the same 50-task training set with identical configurations, varying only the evolver backbone: Claude Sonnet~4.5 vs.\ Claude Haiku~4.5. The solver remains fixed as Claude Haiku~4.5 in both conditions.

\vspace{0.5em}
\noindent\textbf{Observation 1: diagnosis depth and artifact quality.}
We observe stark qualitative differences in how evolvers diagnose failures and synthesize artifacts. Consider a recurring failure pattern: the solver successfully authenticates and retrieves playlists but fails to find the ``most-liked song'' due to incorrect interpretation of the \texttt{like\_count} field versus \texttt{popularity} score.

\begin{itemize}[leftmargin=*]
    \item \textbf{Sonnet 4.5 Evolver:} The diagnosis identifies the \emph{structural} root cause: ``\textit{The agent's technical execution is flawless ($40\%$ score suggests partial credit for correct workflow), but it's missing the actual task requirements. The trajectory shows it reads `What is the title of...' but does not extract the semantic meaning.}'' The evolver proposes a multi-artifact update:
    \begin{itemize}
        \item \texttt{EVOLVE\_TOOL}: Updates \texttt{analyze\_task\_requirements} to parse task descriptions and extract expected output formats
        \item \texttt{ADD\_SKILL}: Creates \texttt{pre\_submission\_verification} to validate answers against task requirements before calling \texttt{task\_complete()}
        \item \texttt{ADD\_SKILL}: Creates \texttt{detect\_execution\_stall} to recognize when the agent is not progressing toward the goal
    \end{itemize}
    % All artifacts pass verification and are committed. By batch 9, the evolved agent has accumulated 4 tools, 16 skills, and 4 knowledge entries.
    
    \item \textbf{Haiku 4.5 Evolver:} The diagnosis is shallower: ``\textit{The agent failed because it did not call the correct API. Fix by adding a tactical patch.}'' The evolver proposes:
    \begin{enumerate}
        \item \texttt{EVOLVE\_TOOL}: Attempts to create \texttt{api\_interaction\_engine} to wrap all API calls
    \end{enumerate}
    In addition, the tool fails verification after 3 repair attempts due to missing parameter validation. The structural tool is rejected, leaving only shallow behavioral textual patches. 
    % By the equivalent batch, the Haiku-evolved agent has only 1 (partially functional) tool, 10 skills, and 5 knowledge entries.
\end{itemize}

\vspace{0.5em}
\noindent\textbf{Observation 2: verification failure rates.}
We quantify the difference in artifact quality by examining verification outcomes:

\begin{table}[h]
\centering
\small
\caption{Verification outcomes across 50-task training runs. Larger evolvers produce artifacts that pass verification at significantly higher rates.}
\begin{tabular}{lccc}
\toprule
\textbf{Evolver} & \textbf{Proposals} & \textbf{Tools Committed} & \textbf{Verification Failures} \\
\midrule
Sonnet 4.5 & 21 & 4 & 2 (repaired successfully) \\
Haiku 4.5 & 25 & 1 & 8 (3+ repair attempts each) \\
\bottomrule
\end{tabular}
\label{tab:evolver_verification}
\end{table}

The Haiku evolver emits $4\times$ more verification failures despite generating fewer total proposals. Qualitative inspection reveals that Haiku-generated tools often have subtle bugs: missing required parameters, incorrect function signatures, or hallucinated API endpoints.

\vspace{0.5em}
\noindent\textbf{Observation 3: cascading effects on solver behavior.}
The quality gap compounds over training. With Sonnet-evolved artifacts, the solver learns to:
\begin{itemize}[leftmargin=*]
    \item Call \texttt{analyze\_task\_requirements()} before execution to parse the goal
    \item Invoke evolved authentication tools that correctly manage access tokens
    \item Apply \texttt{pre\_submission\_verification} to catch answer-format mismatches
\end{itemize}
However, with Haiku-evolved artifacts, the solver exhibits \emph{regression}: it attempts to use the partially-broken \texttt{api\_interaction\_engine} tool, which fails silently or returns errors. The solver then falls back, losing any efficiency gains from tool use.

\vspace{0.5em}
\noindent\textbf{Observation 4: trajectory comparison.}
For task ``venmo payment request based on SimpleNote records'', we compare solver behavior:

\begin{itemize}[leftmargin=*]
    \item \textbf{Sonnet-evolved solver:} Invokes \texttt{analyze\_task\_requirements()} on first turn, retrieves relevant skills, and completes the task in 8 steps with no errors.
    \item \textbf{Haiku-evolved solver:} Fail to call tools, it then attempts 14 exploratory API calls before arriving at the correct workflow. Despite executing the business logic correctly (e.g., creating 4 payment requests), it fails to call \texttt{task\_complete()}, a failure mode that triggered repeated but unsuccessful evolution attempts.
\end{itemize}

\vspace{0.5em}
\noindent\textbf{Takeaway.}
Larger evolvers produce higher-quality diagnoses that identify \emph{structural} capability gaps rather than surface-level symptoms. This translates into more robust artifact synthesis: tools that pass verification, skills that encode generalizable procedures, and knowledge that captures system invariants. The verification stage acts as a quality filter, but cannot rescue fundamentally flawed proposals---hence the importance of evolver capacity in converting $C_{\mathrm{evolve}}$ into durable capability.

\end{document}